\documentclass{article}

\usepackage{microtype}
\usepackage{graphicx}
\usepackage{subfigure}
\usepackage{booktabs} 

\usepackage{hyperref}

\usepackage[accepted]{icml2025}

\usepackage{amsmath}
\usepackage{amssymb}
\usepackage{mathtools}
\usepackage{amsthm}

\usepackage{multirow}
\usepackage{makecell}

\usepackage[capitalize,noabbrev]{cleveref}

\usepackage{url}

\usepackage{algorithm}
\usepackage{algorithmic}

\usepackage{booktabs}
\usepackage{multirow}
\usepackage{multicol}
\usepackage{makecell}
\usepackage{diagbox}
\usepackage{amsmath}
\usepackage{amssymb}
\usepackage{amsfonts} 
\usepackage{xcolor, soul, color}
\usepackage{tabularx}
\usepackage{anyfontsize}
\usepackage{float}
\usepackage{bbm}
\usepackage{dsfont}
\def\mathbi#1{\textbf{\em #1}} 

\usepackage{caption}
\usepackage{graphicx}
\usepackage{lipsum}
\usepackage{wrapfig}

\usepackage{listings}
\usepackage{xcolor}
\usepackage{mathtools}

\theoremstyle{plain}

\theoremstyle{definition}

\theoremstyle{remark}

\usepackage[textsize=tiny]{todonotes}

\usepackage{xspace}
\def\ie{\emph{i.e}\@\xspace} 
\def\ie{\emph{i.e}\@\xspace}

\icmltitlerunning{Under Review}

\begin{document}

\twocolumn[
\icmltitle{Informed Mixing -- Improving Open Set Recognition via \\ Attribution-based Augmentation}



\icmlsetsymbol{equal}{*}

\begin{icmlauthorlist}
\icmlauthor{Jiawen Xu}{berlin}
\icmlauthor{Odej Kao}{berlin}
\icmlauthor{Margret Keuper}{mannheim}

\end{icmlauthorlist}

\icmlaffiliation{berlin}{Distributed and Operating Systems Group, Technical University Berlin, Berlin, Germany}
\icmlaffiliation{mannheim}{Data and Web Science Group, Mannheim University, Mannheim, Germany}

\icmlcorrespondingauthor{Margret Keuper}{keuper@uni-mannheim.de}


\vskip 0.3in
]



\printAffiliationsAndNotice{}  

\begin{abstract}
Open set recognition (OSR) is devised to address the problem of detecting novel classes during model inference. Even in recent vision models, this remains an open issue which is receiving increasing attention. Thereby, a crucial challenge is to learn features that are relevant for unseen categories from given data, for which these features might not be discriminative.
To facilitate this process and "optimize to learn" more diverse features, we propose \textit{GradMix}, a data augmentation method that dynamically leverages gradient-based attribution maps of the model during training to mask out already learned concepts. Thus GradMix encourages the model to learn a more complete set of representative features from the same data source. 
Extensive experiments on open set recognition, close set classification, and out-of-distribution detection reveal that our method can often outperform the state-of-the-art. GradMix can further increase model robustness to corruptions as well as downstream classification performance for self-supervised learning, indicating its benefit for model generalization. 
\end{abstract}

\section{Introduction}
\label{sec-introduction}

Deep neural networks have achieved remarkable performance across various fields, particularly in object classification tasks \cite{russakovsky2015imagenet}. However, traditional learning paradigms require the entire dataset to be available prior to training, and models are typically limited to recognizing the classes present in the training set. In real-world scenarios, however, new and previously unseen classes often emerge, which may be difficult to collect in the short term or even impossible to anticipate beforehand. Hence, the challenge of recognizing novel classes during inference becomes unavoidable. This task is known as open set recognition (OSR), and it has garnered significant research attention \cite{asg_Yu17, GOpenMax_Ge17, hassen2020learning, dhamija2018reducing, yoshihashi2019classification, zhou2021learning, cao2021open, miller2021class, arpl, vaze2022openset, xu2023contrastive, wang2024exploring}. 
Throughout this paper, we denote known classes using "in set" and unknown classes with "open set".

\begin{figure}
    \centering
    \includegraphics[width=0.9\linewidth]{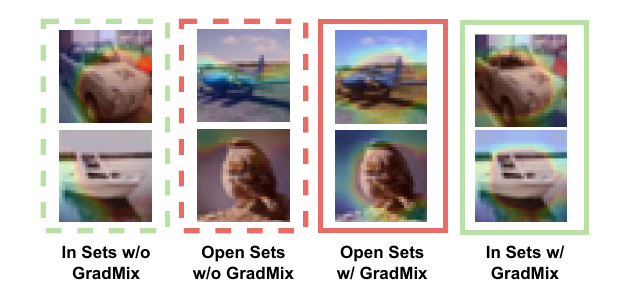}
    \caption{An illustration of GradMix’s effectiveness using attribution maps (green and red boxes for in set (car, boat) and open set (airplane, bird) samples respectively; dashed-line boxes indicate models without GradMix, while solid-line boxes represent models with GradMix). GradMix enables models to focus on broader areas in data for in set samples while capturing object regions more effectively for open set samples.}
    \label{fig-treser}
\end{figure}

Existing OSR approaches leverage a range of techniques, including generative models \cite{asg_Yu17, GOpenMax_Ge17}, novel feature learning objectives \cite{miller2021class}, and ensembling \cite{wang2024exploring}. Most of these methods are fundamentally based on the principle of comprehensively modeling the in set data. Theoretical work in \cite{wang2024exploring} has demonstrated that OSR performance is positively correlated with feature diversity. 
In this paper, we propose to enhance models' ability of learning more diverse features from in set data through two key aspects.
First, in conventional supervised learning paradigms, the phenomenon of class collapse, where sub-class features are suppressed \cite{jing2021understanding, xue2023features, chen2022perfectly}, can hinder the learning of diverse features. By combining supervised and self-supervised learning, this issue can be well mitigated \cite{chen2022perfectly}.

\begin{table}[ht]
\centering
\fontsize{6pt}{6pt}\selectfont
\begin{tabular}{c c c c c c} 
 \toprule
 Methods & Datasets & ID & $A_\textit{nno}$ & $A_\textit{enp}$ & $A_\textit{msp}$   \\[1mm]
 \midrule
 \multirow{ 2}{*}{CE}              &CIFAR10       &  $16.68$    & $74.11$   &   $80.76$  &  $82.83$ \\
                                    &TinyImageNet  &  $15.54$   &  $68.85$ & $66.71$  &  $67.5$ \\ 
 \multirow{ 2}{*}{\makecell{CE \\+ SSL}} & CIFAR10 &  $\mathbf{17.15}$   &  $\mathbf{83.68}$  &   $\mathbf{84.49}$ &  $\mathbf{85.11}$ \\
     &  TinyImageNet  & $\mathbf{20}$   &  $\mathbf{74.58}$ &   $\mathbf{72.53}$ &  $\mathbf{72.75}$ \\[3mm]
     
 \multirow{ 2}{*}{SupCon} &  CIFAR10 &   $8.07$   &   $67.15$  & - & -    \\
                              &  TinyImageNet  & $10.04$  &  $72.97$ & - & - \\ 
 \multirow{ 2}{*}{\makecell{SupCon \\+ SSL}} & CIFAR10 &    $\mathbf{14.25}$ & $\mathbf{88.75}$ & - & -   \\
     &  TinyImageNet  &  $\mathbf{17.47}$  &  $\mathbf{77.48}$ & - & -  \\
     
\bottomrule
\end{tabular}
\caption{A comparison between supervised learning models trained with and without SSL is made based on ID and AUROC (higher ID indicates more diverse features). Results in each group showing higher feature diversity and better OSR performance are highlighted in bold. The symbol “-” indicates that the specified detection method is not applicable to the corresponding model. Results indicate that higher feature diversity coindice with improved OSR performance.
}
\label{tab-div-osr}
\end{table}

For this, we conduct concise experiments to assess OSR performance and measure feature diversity using the Intrinsic Dimension (ID) \cite{ansuini2019intrinsic}. 
Two groups of models are trained under supervised learning paradigms: cross-entropy loss (CE) and supervised contrastive learning (SupCon) \cite{khosla2020supervised}. In each group, we evaluate models trained with and without self-supervised contrastive learning (SSL) \cite{chen2020simple}, resulting in four sets of learning objectives: CE, CE+SSL, SupCon, and SupCon+SSL.
We use the CIFAR10 and TinyImageNet protocols to evaluate OSR performance, as described in Section \ref{subsec-exp-osr}.
For CE and CE+SSL models, we apply three methods: the entropy of output logits, the maximum output logit, and the Nearest Non-Outlier (NNO) \cite{scheirer2012toward} to compute scores for detecting open set samples. For SupCon and SupCon+SSL, NNO is applied due to a lack of a classification head.
We denote the AUROCs calculated using the above scores using $A_\textit{nno}$, $A_\textit{enp}$, and $A_\textit{msp}$, respectively. 
More on the experimental settings is in Appendix \ref{app-introduction-exp}. The results in Table \ref{tab-div-osr} demonstrate that models trained with SSL learn more diverse features and achieve superior OSR performances, which can also empirically verify the conclusion in \cite{wang2024exploring}.
Based on these findings, we adopt this approach for OSR.

Furthermore, we propose a novel data augmentation technique to enhance feature diversity by allowing the models to dynamically focus on previously unlearned areas within the data. Leveraging LayerCAM \cite{jiang2021layercam}, a gradient-based method originally developed for visual explanations, we identify the regions that are most activated during training. We then mask these highly activated areas to generate augmented data. This technique, which we refer to as \textbf{GradMix}, has shown improved OSR performances compared to other popular mixing data augmentation methods. As illustrated in Figure \ref{fig-treser}, models (SupCon+SSL) trained with GradMix can focus on broader areas in data for in sets and capture objects in open set data more effectively.
Moreover, GradMix has proven advantageous for classification under common corruptions and for improving downstream classification accuracy in self-supervised learning.

Our contributions are as follows:
\begin{enumerate}
     \item We apply supervised and self-supervised contrastive learning for OSR, and empirically demonstrate its effectiveness in promoting diverse feature learning.  \label{item-1}
     \item We have designed a novel data augmentation method, GradMix, for \ref{item-1}, leveraging attribution information during training. Extensive experiments show that our OSR method achieves state-of-the-art performance.
    \item Besides, GradMix improves classification performance under common corruptions and the downstream classification accuracy in self-supervised learning on diverse network architectures and datasets. 
\end{enumerate}

\section{Related Work}
\label{sec-relatedwork}

\subsection{Open Set Recognition}  \label{subsec-relatedwork-osr}
Open set recognition (OSR) involves identifying novel class samples during inference. OSR methods generally fall into two categories: discriminative and generative models. Discriminative models \cite{asg_Yu17, miller2021class, hassen2020learning, arpl, vaze2022openset, xu2023contrastive, wang2024exploring} can either focus on designing learning objectives that enhance feature learning for known classes, such as \cite{miller2021class}, or leverage synthesized open set samples generated by generative models like GANs to train a $C+1$ ($C$ is the number of close set classes) classifier, such as \cite{asg_Yu17}.


Another category utilizes generative models to model in set data \cite{GOpenMax_Ge17, cao2021open}. For instance, \cite{cao2021open} utilizes a Gaussian mixture variational autoencoder to model in set data. For a more thorough review of open set recognition methods, see \cite{mahdavi2021survey}.

\subsection{Mixing Data Augmentation}  \label{subsec-relatedwork-mixing}
Mixing data augmentation combines raw samples with other images from within or outside the dataset. These methods can be broadly categorized into two types.
The first category is pixel-wise mixing, initially introduced as \textit{Mixup} \cite{zhang2017mixup}, in which augmented data is created by taking a weighted average of the original samples and their corresponding labels. 

The second category is patch-wise mixing, where a portion of the original samples is replaced. An early method in this category is \textit{CutOut} \cite{devries2017improved}, which involves masking out a patch of the original samples. \textit{CutMix} \cite{yun2019cutmix} extends this idea by replacing patches with resized samples from the same minibatch. 
Besides, there are many other mixing data augmentation approaches. \cite{lewy2023overview} presents a comprehensive review.

\section{Method} \label{sec-method}

Our approach is centered on the concept of learning diverse features through two primary components: the incorporation of self-supervised learning within supervised paradigms and the application of the proposed gradient-based mixing data augmentation method, \emph{GradMix}. 
For clarity in the following text, we adopt the following notation: uppercase letters represent sets of scalers, while lowercase letters refer to individual samples within those sets. Vectors are indicated by bold letters, and uppercase bold letters are used to denote sets of vectors or matrices.

\subsection{Supervised and Self-supervised Contrastive Learning} \label{sec-diverse-feature-learning}

Self-supervised learning has recently gained substantial attention for its ability in extracting high-quality features without the need for labeled data \cite{chen2020simple, he2020momentum, grill2020bootstrap}. As demonstrated in Section \ref{sec-introduction}, combining supervised and self-supervised learning can boost OSR performance.
We start with this strategy in our open set recognition framework.

We pair SimCLR \cite{chen2020simple} and supervised contrastive learning (SupCon) \cite{khosla2020supervised} in our method due to their shared foundational and superior performance in OSR as shown in Table \ref{tab-div-osr} compared with cross-entropy models.
SimCLR is a self-supervised contrastive learning method that learns representations by minimizing the distances between the original data and its augmented counterpart in the feature space. For a sample $\mathbf{x}_i$ and its augmented version $\mathbf{x}_j$ within a minibatch of size $2N$ (with $N$ original samples and their corresponding augmentations), their representations are denoted as $\mathbf{z}_i$ and $\mathbf{z}_j$. $\mathbf{x}_i$ and $\mathbf{x}_j$ form a positive pair, while $\mathbf{x}_i$ and all other samples in the minibatch constitute negative pairs.
SimCLR is designed to maximize the mutual information between $\mathbf{z}_i$ and $\mathbf{z}_j$ \cite{oord2018representation}, encouraging the model to capture more of the inherent features in the sample itself as described in ~\eqref{equ-simclr}.
In ~\eqref{equ-simclr}, $\mathbbm{1}_{[k \neq i]} \in {0, 1}$ is an indicator function that flags negative pairs, \ie, $\mathbf{x}_i$ paired with any samples other than $\mathbf{x}_j$ in the minibatch. The function $sim(\cdot)$ measures similarity, typically using the cosine function. The temperature parameter $\tau$ is a hyperparameter that controls the learning dynamics between positive and negative pairs.
It is clear from ~\eqref{equ-simclr} that minimizing $L_{\textit{simclr}}(i)$ is equivalent to maximizing $\textit{sim}(\mathbf{z_i}, \mathbf{z_j})$ while minimizing $\textit{sim}(\mathbf{z_i}, \mathbf{z_k})$. 

\begin{equation}
\small
L_{\textit{simclr}}(i) = -\log \frac{\exp (\textit{sim}(\mathbf{z_i}, \mathbf{z_j}) / \tau)}{\sum_{k=1}^{2N} \mathbbm{1}_{[k \neq i]} \exp (\textit{sim}(\mathbf{z_i}, \mathbf{z_k}) / \tau)} 
    \label{equ-simclr}
\end{equation}
SupCon extends SimCLR to a supervised fashion and has been shown to outperform cross-entropy loss in terms of model generalization \cite{khosla2020supervised}. In SupCon, positive pairs are samples belonging to the same class, while negative pairs consist of samples from different classes. Here, labels are denoted by $\ell$.
The set of all positive pairs for $\mathbf{x}_i$ is denoted by $P(i)$, where $P(i) = \{ {x}_p | 1 \leq p \leq 2N, p \neq i, \ell_i = \ell_p\}$. 
As with SimCLR, the learning objective is to minimize the similarities between the representations of positive pairs while maximize the distances between negative pairs, as described in ~\eqref{equ-supcon}.
 %
\begin{equation}
\small
\begin{split}
    & L_{\textit{supcon}}(i) =   \\
    & \frac{1}{|P(i)|} \sum\limits_{p \in P(i)} -\log \frac{\exp \left({\textit{sim}\left( \textbf{z}_i \cdot \textbf{z}_{p}\right)}/{\tau} \right)}{\sum_{k=1}^{2N} \mathbbm{1}_{l_i \neq l_k} \exp \left({\textit{sim}( \textbf{z}_i \cdot \textbf{z}_k)}/{\tau} \right)}
    \label{equ-supcon}
\end{split}
\end{equation}

We combine $L_{\textit{simclr}}$ and $L_{\textit{supcon}}$ linearly with weights $\theta$ and $\lambda$ respectively in a multi-task learning fashion to minimize class collapse, \ie:
\begin{equation}
\small
    L_{\textit{contra}} = \theta \cdot L_{\textit{supcon}} + \lambda \cdot L_{\textit{simclr}}
\label{equ-cl}
\end{equation}

\subsection{Gradient-based Mixing Augmentation}  \label{subsec-method-gradmix}

To enable the models to learn more diverse features, we developed a data augmentation method that encourages the models to focus on broader areas within the data. Inspired by the mixing augmentation techniques introduced in \ref{subsec-relatedwork-mixing}, our approach masks out a portion of the learned areas after each epoch during training.
Rather than random masking, we expect the model can pay more attention to unlearned areas in data. To achieve this, we propose detecting activated areas in the input data using the models during training, with the help of visual explanation techniques.

One foundational method for visual explanation is GradCAM \cite{selvaraju2017grad}, which visualizes the learning process by performing a weighted combination of the forward feature maps. We denote the attribution map of the activated areas computed by GradCAM as $\mathbf{M}_{gradcam}$. As shown in ~\eqref{equ-gradcam},  $\mathbf{M}_{gradcam}$ is the ReLU-filtered weighted sum of the forward feature maps $\mathbf{A}^k$, where $k$ represents the index of the feature maps in the selected convolutional layer.
As expressed in ~\eqref{equ-gradcam} and ~\eqref{equ-ak}, the partial gradient of the loss function $L$ with respect to the feature map $\mathbf{A}^k$, \ie $ \alpha_k$, indicates the importance of $\mathbf{A}^k$ to $\mathbf{M}_{gradcam}$. 

\begin{equation}
\small
    \mathbf{M}_{\textit{gradcam}} = ReLU(\sum_k \alpha_k \mathbf{A}^k) \quad\text{with}
    \label{equ-gradcam}
\end{equation}
%
\begin{equation}
\small
    \alpha_k = \frac{1}{|\mathbf{A}^k |} \sum_i \sum_j \alpha_{i,j}^k = \frac{1}{|\mathbf{A}^k |} \sum_i \sum_j \frac{\partial L}{\partial \mathbf{A}_{i,j}^k}.
    \label{equ-ak}
\end{equation}

\begin{figure*}
\begin{minipage}[t]{0.68\textwidth}
  \includegraphics[width=\linewidth]{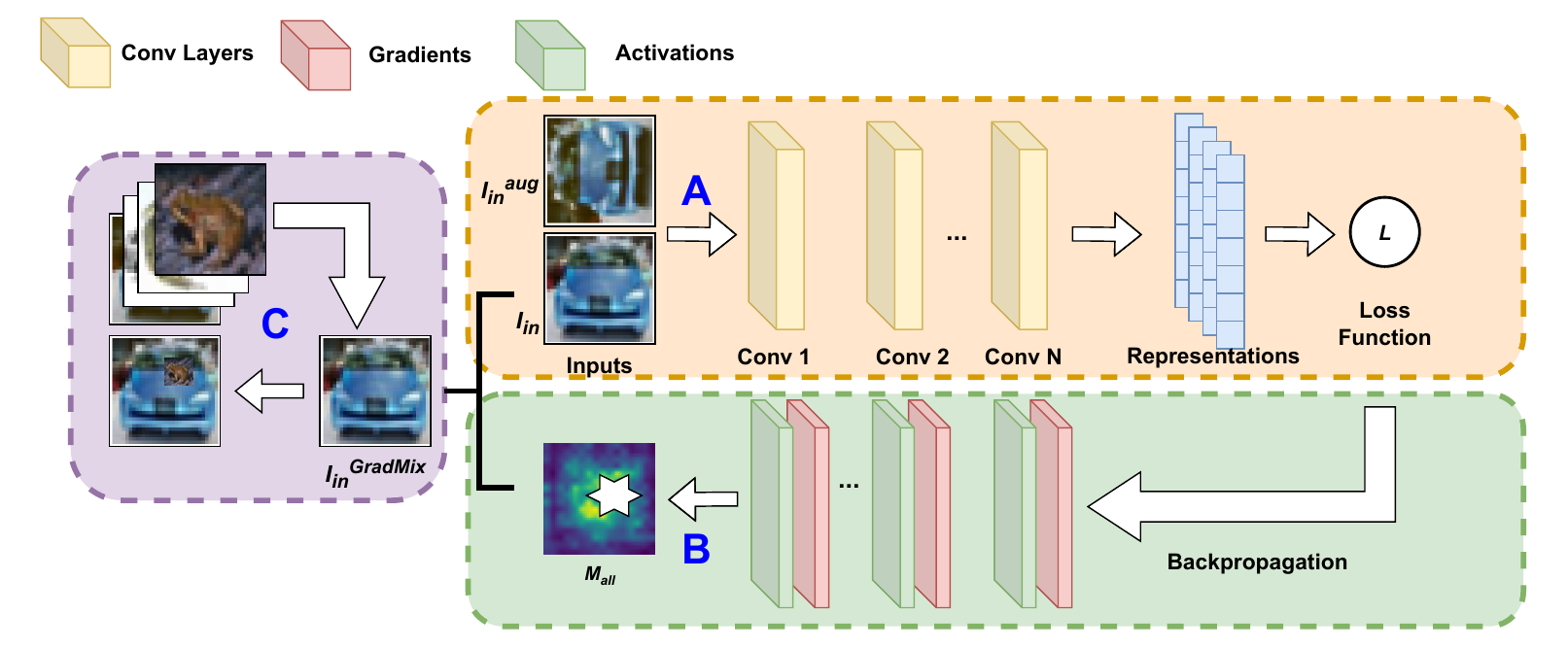}
  \caption{Graphical illustration of GradMix. Three blocks with dashed borderlines refer to the procedures: \textbf{A}. Data is fed into the feature extractor during forward propagation; \textbf{B}: Attribution maps are computed using the internal feature maps and LayerCAM method. The most activated area is selected (highlighted using a white star in the graph); \textbf{C}: A random sample from the same minibatch is resized and patched on the most activated area.}
  \label{fig-method}
\end{minipage}%
\hfill 
\begin{minipage}[t]{0.3\textwidth}
  \includegraphics[width=\linewidth]{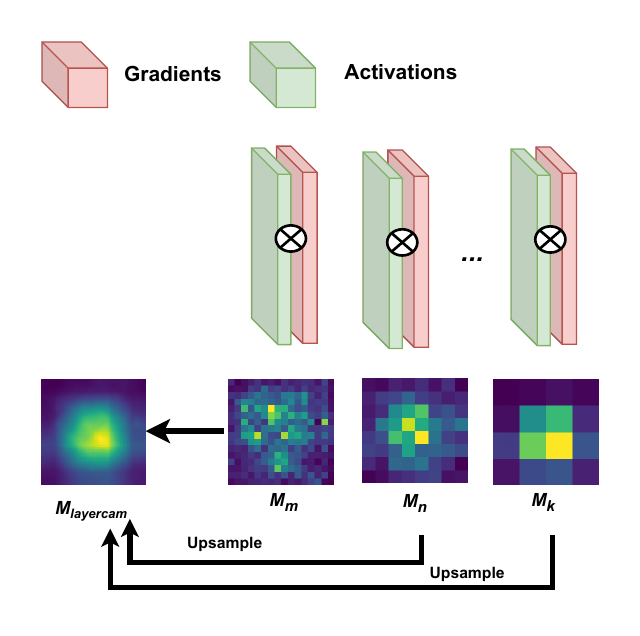}
  \caption{Graphical illustration of the proposed layer aggregation for GradMix: the attribution maps calculated using individual layers are summed to get $\mathbf{M}_\textit{layercam}$.}
  \label{fig-layer-aggregation}
\end{minipage}
\end{figure*}

However, the attribution maps computed using GradCAM tend to be coarse and lack precision in detecting activation areas in the input. To address this, we apply an improved method, LayerCAM \cite{jiang2021layercam}.
Unlike GradCAM, LayerCAM can leverage earlier layers in CNNs to capture finer-grained attribution maps. Instead of using a single weight coefficient, $ \alpha_k$, for the entire feature map channel $\mathbf{A}^k$, as in ~\eqref{equ-gradcam} and ~\eqref{equ-ak}, LayerCAM assigns individual weights to each location in $\mathbf{A}^k$, as in ~\eqref{eq-ak-layer}. The weighted feature maps $\mathbf{\tilde{A}}^k$ are then summed along the channel dimension to produce the final attribution map $\mathbf{M}_\textit{layercam}$: 
\begin{equation}
\small
    \mathbf{\tilde{A}}^k = ReLU(\frac{\partial L}{\partial \mathbf{A}_{i,j}^k}) \odot \mathbf{A}_{i,j}^k \qquad
    \label{eq-ak-layer}
\end{equation}
\begin{equation}
\small
    \mathbf{M}_\textit{layercam} = ReLU\left( \sum_k \mathbf{\tilde{A}}^k\right).
    \label{eq-layer}
\end{equation}
The attribution maps are then used for data augmentation. Figure \ref{fig-method} provides a graphical illustration of our method. The ratio between the side lengths of the masked area and the original image, denoted as $\gamma$, follows a uniform distribution, \ie, $\gamma \sim \mathbf{U}(\gamma_{min}, \gamma_{max})$. In our work, we set $\gamma_{min}$ to $0.1$ and $\gamma_{max}$ to $0.5$, ensuring that the masks can neither completely cover the entire object in the original samples nor exceed the image margins. We name our method \emph{GradMix}.
\emph{GradMix} is applied to the self-supervised component in \eqref{equ-cl}, and its loss values are weighted by the ratio of the masked area to the original image, which is $\gamma^2$. The overall learning objective then becomes:
\begin{equation}
\small
    L = \theta \cdot L_{\textit{supcon}} + \lambda \cdot (L_{\textit{simclr}} + \gamma^2 \cdot L_{\textit{simclr}}^{\textit{GradMix}})
\label{equ-loss-all}
\end{equation}
Furthermore, as shown in \ref{osr-ablation-layers}, we found that aggregating attribution maps from multiple convolutional layers enhances OSR performance. 
Figure \ref{fig-layer-aggregation} illustrates this aggregation process.
The final attribution map is the sum of these computed using individual convolutional layers, \ie, $\mathbf{M}_\textit{layercam} = \mathbf{M}_{m} + \mathbf{M}_{n} + ...+ \mathbf{M}_{k}$, $m$, $n$, and $k$ are selected indices of the convolutional layers.

A prior work, Attentive CutMix, in \cite{walawalkar2020attentive}, utilizes attention maps generated from individual pre-trained models to localize image patches that to be mixed with other data. However, we argue that Attentive CutMix and \emph{GradMix} originate from fundamentally different principles and are based on distinct conceptual approaches. Additionally, \emph{GradMix} can demonstrate far better OSR performance as shown in \cref{subsec-exp-osr}.
We provide a detailed comparison in Appendix \ref{app-compare}.

\subsection{OSR Framework} \label{subsec-method-osr-framework}

We apply a distance-based method to detect open set samples. Our method is close to Nearest
Non-Outlier (NNO) \cite{scheirer2012toward} with the idea that if a testing sample is far from all in set classes then it belongs to the open set. However, unlike NNO which uses the class mean to represent each class, our method computes the sum of the distances, $d_{i}^{c}$, between the testing sample $\mathbf{x}_i$ and $K$ closest samples of each class $c$ in feature space. The distances are measured using cosine similarity. The maximum of the normalized $d_{i}^{c}$ serves as the score, $s_{i}$, to detect open sets, \ie, $s_{i} = \max_{c}\frac{d_{i}^{c}}{\sum_c d_{i}^{c}}$. Appendix \ref{app-detection-method} provides more details.

\section{Experiments} \label{sec-exp}

In this section, we present the experimental settings and results for open set recognition, close set classification, out-of-distribution detection, corrupted image classification, as well as model generalization. 

\subsection{Open Set Recognition} \label{subsec-exp-osr}

\paragraph{Settings  }
Following the OSR testbench widely used in the literature \cite{arpl, vaze2022openset, neal2018open}, we evaluate our method on six open and close set split protocols: MNIST, SVHN, CIFAR10, CIFAR+10, CIFAR+50, and TinyImageNet, which are created using the source datasets MNIST \cite{deng2012mnist}, SVHN \cite{Netzer2011SVHN}, CIFAR10 \cite{cifar10}, CIFAR100 \cite{cifar100}, and TinyImageNet \cite{deng2009imagenet}. In each protocol, five different splits are applied and all final results are the average of these five trials. 
ResNet18 \cite{he2016deep} is used as the feature encoder backbone in all experiments of this section, with the output feature dimension set to 128. 
We use the area under the receiver operating characteristics curve (AUROC) to evaluate the OSR performance, which is the most common metric for OSR and can be interpreted as how much the detection score histograms of open set and in set samples are overlapped.
AUROC is threshold-independent and higher AUROC represents better performance at detecting open set samples. The complete experimental settings are detailed in Appendix \ref{app-osr-exp}.

\paragraph{Results  }
We compare our results with vanilla cross entropy classifier, as well as the state-of-the-art and popular OSR methods, namely Openmax \cite{bendale2016towards}, G-Openmax  \cite{GOpenMax_Ge17}, OSRCI \cite{neal2018open}, C2AE \cite{oza2019c2ae}, GRROSR \cite{perera2020generative}, PROSER \cite{zhou2021learning}, APRL \cite{arpl}, APRL-CS \cite{arpl}, OpenAUC \cite{wang2022openauc}, ConOSR \cite{xu2023contrastive}, and MEDAF \cite{wang2024exploring}.
Additionally, we test Attentive CutMix with its settings in the original paper \cite{walawalkar2020attentive}.
The results are presented in Table \ref{tab-auroc-results}. 
Our method achieves the best performance on most protocols, especially complex protocols, such as CIFAR+50 and TinyImageNet.
Particularly, it demonstrates a clear advantage on TinyImageNet with over $1\%$ of increase. We think the reasons lie in the complexity of the dataset. The performance of contrastive learning can be increased when harder and more variant negative samples are introduced during training \cite{shu2024unsupervised}. And complex data can provide larger room for GradMix to mine more features.

\begin{table}[ht]
\begin{center}
\fontsize{6pt}{6pt}\selectfont
\begin{tabular*}{\columnwidth}{l @{\,\,\,} c@{\,\,\,\,}c@{\,\,\,}c@{\,}c@{}c@{}c@{\,}}
 \toprule
 &  \rotatebox{30}{{MNIST}} & \rotatebox{30}{{SVHN}} & \rotatebox{30}{{CIFAR10}} & \rotatebox{30}{{CIFAR+10}} & \rotatebox{30}{{CIFAR+50}}  & \rotatebox{30}{{TinyImgNet}} \\ 
 \midrule
 Cross Entropy &  $97.8 $  &  $88.6 $  &  $67.7 $  & $81.6 $  & $80.5 $   &  $ 57.7 $  \\
 Openmax  \cite{bendale2016towards} &  $98.1 $  &  $89.4$  &  $69.5 $  & $81.7 $  & $79.6$   &  $ 57.6 $  \\
 G-Openmax  \cite{GOpenMax_Ge17}    &   $98.4 $  &  $89.6 $  &  $67.5$  & $82.7$  & $81.9$   &  $ 58.0$  \\
 OSRCI  \cite{neal2018open}  &  $98.8$  &  $90.1$  &  $69.9$  & $83.8$  & $82.7$   &  $ 58.6 $  \\
 C2AE \cite{oza2019c2ae}      & $98.9 $  &  $92.2$  &  $89.5 $  & $95.5$  & $93.7$   &  $ 74.8$  \\
 GRROSR \cite{perera2020generative}  & - &  $93.5$  &  $80.7 $  & $92.8$  & $92.6$   &  $ 60.8$ \\
 PROSER \cite{zhou2021learning} & - &  $94.3$  &  $89.1 $  & $96.0$  & $95.3$   &  $ 69.3$ \\
 APRL \cite{arpl}            & $99.6$  &  $96.3$  &  $90.1$  & $96.5$  & $94.3$   &  $ 76.2$  \\
 APRL-CS \cite{arpl}         & $99.7$  &  $96.7$  &  $91.0 $  & $97.1$  & $95.1 $   &  $ 78.2$  \\
 OpenAUC \cite{wang2022openauc} & $99.4$   & $95.0$  & $89.2$  & $95.2$ & $93.6$  &  $75.9$   \\ 
 ConOSR \cite{xu2023contrastive} &  $99.7$   & $\mathbf{99.1}$  & $\mathbf{94.2}$  & $98.1$ & $97.3$  &  $80.9$   \\ 
 MEDAF \cite{wang2024exploring} & -   & $95.7$  & $86$  & $96$ & $95.5$  &  $80.0$    \\
 [0.5ex] 
 \hline
 Attentive CutMix          & $78.75$   & $79.91$  &  $74.32$  & $84.51$  & $89.65$  &  $64.52$   \\
\hline
GradMix (Ours)    &   $\mathbf{99.8}$ &  $94.7$  & $91.33$  & $\mathbf{98.62}$ & $\mathbf{97.64}$ &   $\mathbf{81.92}$ \\ 

 \bottomrule
\end{tabular*}
\caption{The area under the ROC curve (AUROC) (in $\%$) for detecting known and unknown samples (Partial results of the baseline methods are from \cite{arpl} and \cite{xu2023contrastive}). $"-"$ indicates there are no given results in the literature.
Bold numbers indicate the best results. GradMix outperforms the SoTAs in four out of six protocols and the increase is over $1\%$ on large-scale TinyImageNet protocol.\label{tab-auroc-results}}
\label{tab-auroc-results}
\end{center}
\end{table}

\paragraph{Ablations} 
We perform two ablation studies: (1) comparing OSR performance when applying different mixing data augmentation techniques, and (2) assessing the impact of utilizing different deep layers as well as layer aggregation introduced in \ref{subsec-method-gradmix} for computing attribution maps in GradMix. Consistent with previous experiments, OSR performance is evaluated using AUROC and each result is the average of five trials. All experiments are repeated with CIFAR10 and TinyImageNet protocols. We fix $k=10$ (see \ref{subsec-method-osr-framework}) in each study for fair comparison.

\subparagraph{A. Impact of Different Augmentation Methods} We have (re)-implemented and tested three augmentation methods \ie, Mixup \cite{zhang2017mixup}, CutMix \cite{yun2019cutmix}, and GradMix, as well as the models without extra augmentation. Furthermore, we vary the hyper-parameter $\alpha$ in Mixup and CutMix. Larger $\alpha$ leads to stronger augmentations (see \cite{zhang2017mixup, yun2019cutmix} for details).   

\begin{figure}[ht]
\centering
  \includegraphics[width=0.9\linewidth]{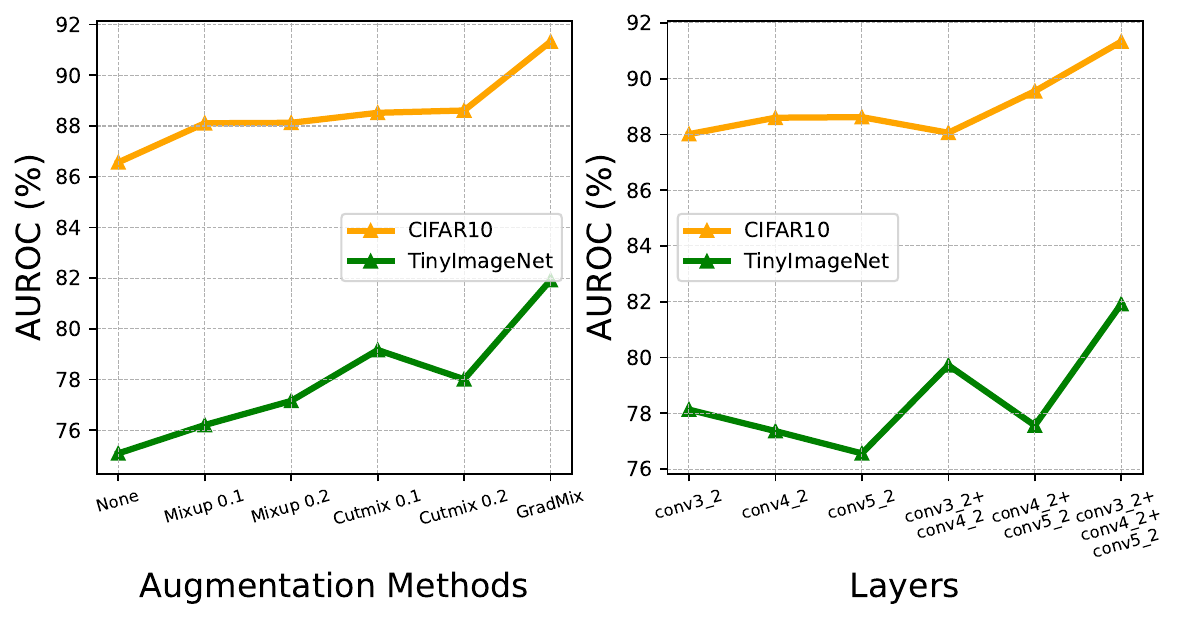}
  \label{fig-ablation_layer}
   \caption{\textbf{Left:} OSR performances of the models with different augmentation methods on CIFAR10 and TinyImageNet protocols. Clear improvements can be brought by extra data augmentations. And GradMix performs best among all augmentation methods. \textbf{Right:} OSR performances of models with GradMix computed using different layers and layer aggregation in ResNet18. The results indicate that different layers can produce non-negligible changes in performance. And layer aggregation can in general provide better performance than single layers. \label{fig-ablation_method}}
\end{figure}

The results, as shown in Figure \ref{fig-ablation_method} (left), clearly demonstrate that the models incorporating augmentations achieve superior OSR performance across both protocols. GradMix, in particular, improves AUROC by over $3\%$ compared to without extra augmentation on both protocols. Furthermore, more advanced and stronger augmentation techniques can result in greater performance gains, with GradMix consistently outperforming all other methods.

\subparagraph{B. Selection of Deep Layers in GradMix} \label{osr-ablation-layers}
In order to investigate the impact of different deep layers and their aggregations used for computing attribution maps in GradMix, we evaluate the OSR performances of the models with GradMix computed using various layers of ResNet18, specifically \textsf{conv3\_2}, \textsf{conv4\_2}, \textsf{conv5\_2}, $\textsf{conv3\_2}+\textsf{conv4\_2}$, $\textsf{conv4\_2}+\textsf{conv5\_2}$, $\textsf{conv3\_2}+\textsf{conv4\_2}+\textsf{conv5\_2}$. The latter three configurations represent layer aggregations. 

The results are summarized in Figure \ref{fig-ablation_method} (right), which indicate that the utilization of different layers, as well as layer aggregation, introduce non-negligible performance changes, especially for higher resolution data, \ie, TinyImageNet. For both testing protocols, layer aggregation brings obvious improvements to OSR performance. For CIFAR10, GradMix with early layers imposes better performance whereas it is the inverse case for TinyImageNet. We think the reason lies in that the features are richer and more distributed for more complex data and attribution maps from early layers can better preserve these features. Based on the above results, we adopt layer aggregation in our method.

\subsection{Close Set Classification} \label{subsec-close-set}

\paragraph{Settings} 
In order to evaluate the proposed method on close set classification tasks, we train models on full CIFAR10, CIFAR100, and half TinyImageNet (the first 100 classes) datasets and test their classification accuracy as in \cite{arpl, xu2023contrastive, wang2024exploring}. We compare the results with cross-entropy, ARPL, ConOSR, and MEDAF.
Results are shown in table \ref{tab-close-set}.

\begin{table}[ht]
   \fontsize{6pt}{6pt}\selectfont
   \centering
    \begin{tabular}{@{}cccc@{}}
     \toprule
         Method & CIFAR10 & CIFAR100 & TinyImgNet  \\[0.5ex]
    \midrule
        Cross-Entropy &  $94$  &   $71.6$   & $63.7$          \\
        ARPL      &  $94.1$  &   $72.1$   & $65.7$          \\
        ConOSR    &  $94.6$  &   $73$   & $66.1$          \\
        MEDAF    &  $95.4$  &   $77$   & $70.6$          \\
    \midrule
       GradMix (Ours)     & $94.1$   & $\mathbf{78}$   &    $\mathbf{72.46}$          \\
    \bottomrule
    \end{tabular}
    \caption{Comparison on close set classification performances (classification accuracy in $\%$) on CIFAR10, CIFAR100, and half TinyImageNet datasets. The results of baselines are from \cite{arpl, xu2023contrastive, wang2024exploring}. Our method can outperform the baselines on the two larger datasets, CIFAR100 and TinyImageNet, with significant increases.} 
    \label{tab-close-set}
\end{table}

\begin{table*}[ht]
\begin{center}
 \fontsize{6pt}{6pt}\selectfont
\begin{tabular*}{0.9\textwidth}{c @{\extracolsep{\fill}} cccccccccc}
 \toprule
 \multirow{2}{*}{Method} &  \multicolumn{5}{c}{In: CIFAR10 \/ Out: CIFAR100}   &   \multicolumn{5}{c}{In: CIFAR10 \/ Out: SVHN}     \\
 \cmidrule(lr){2-6} \cmidrule(rl){7-11}
 & TNR & AUROC  &   DTACC    &  AUIN  &  AUOUT  &                        TNR & AUROC  &   DTACC    &  AUIN  &  AUOUT    \\
 \hline
 Cross Entropy  & $31.9$  &  $86.3$  & $79.8$  &  $88.4$ & $82.5$  &     $32.1$ & $90.6$ &  $86.4$  & $88.3$  &   $93.6$ \\
 ARPL  & $47.0$   &   $89.7$    &  $82.6$  &  $90.5$ & $87.8$  &   $53.8$  &  $93.2$  &  $87.2$   &  $90.3$  &  $95.8$  \\
 APRL-CS & $48.5$   &   $90.3$    &  $83.4$  &  $91.1$ & $88.4$  &   $79.1$  &  $96.6$  &  $91.6$   &  $94.8$  &  $98.0$  \\
 MEDAF   & -  &  $92.5$ &  $85.4$ &  $93.2$ & $91.1$  &   -  &  $99.1$  &  $95.3$  &  $98.0$  & $99.6$  \\ [0.5ex] 
 \hline
 GradMix (Ours)    & $\mathbf{50.3}$  &  91.63  &  $\textbf{92.15}$  & $92.5$  &  $90.86$ &     $\mathbf{80.5}$  & $96.7$  & $\mathbf{96.82}$   &    $96.9$ & $95.3$  \\ [0.5ex] 
 \bottomrule
\end{tabular*}
\caption{GradMix performs comparably or better than previous works in out of distribution detection. Results of baselines are from \cite{arpl, wang2024exploring}.\vspace{-0.5cm}}
\label{tab-ood-results}
\end{center}
\end{table*}
\begin{figure*}
 \begin{minipage}[.1\textheight]{.65\linewidth}
    \includegraphics[width=0.85\linewidth]{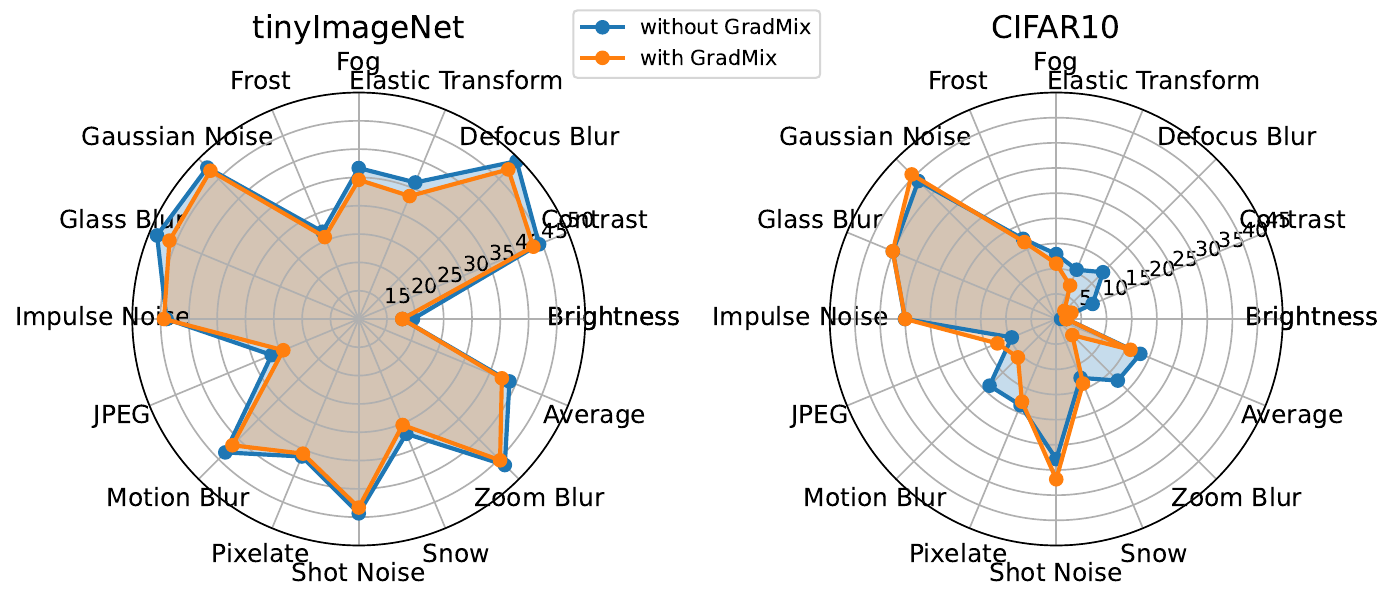}
    \caption{Classification accuracy drop of the models trained with and without GradMix on CIFAR10 and TinyImgNet. The average accuracy drop $\bar D_{c}$s are lower for the models with GradMix for most corruption types on both datasets.}
    \label{fig-acc-corruption}
    \end{minipage}
  \hfill
  \begin{minipage}[.1\textheight]{.33\linewidth}
  \fontsize{6pt}{6pt}\selectfont
    \begin{tabular}{@{}c@{\,}|@{\,}c@{\,}c@{\,}c@{\,}c@{}}
    \toprule
       & \multicolumn{2}{c}{CIFAR10} & \multicolumn{2}{c}{TinyImageNet} \\
      \midrule
      Severity & w/o GradMix & GradMix & w/o GradMix & GradMix\\
      \midrule
       1   &  7.01&\textbf{6.66}   & \textbf{22.61}&25.25       \\
      2   &  11.18&\textbf{10.04}  & 37.02&\textbf{31.49}       \\
      3   &  15.93&\textbf{14.03}  & 39.43&\textbf{38.91}      \\
      4   &  23.01&\textbf{20.49}  & 46.23&\textbf{45.23}       \\
      5   &  33.29&\textbf{28.64}  & 49.31&\textbf{48.91}      \\
      Avg. & 18.08&\textbf{15.97}  & 38.81&\textbf{37.36}       \\
    \bottomrule
    \end{tabular}
    \captionof{table}{Avg.~accuracy drop $\bar D_{s}$ for different corruption severities of the models without (left) and with GradMix (right) on CIFAR10 and TinyImageNet datasets. Better results are in bold. For almost all severity levels, GradMix model shows lower $\bar D_{s}$ and is therefore more robust.}
    \label{tab-acc-corruption}
   
  \end{minipage}
\end{figure*}

\paragraph{Results} All the baselines are supervised learning methods, which are in principle better at close set classification, whereas our method employs self-supervised. For CIFAR10, there exist very confusing in set class pairs in some trails, e.g., deer and horse, which can significantly lower the close set classification accuracy. Our models can learn more non-discriminative features with self-supervised learning, which could in theory worsen the problem.
However, our method can still achieve significantly better performances on CIFAR100 and TinyImageNet and fair results for CIFAR10. Especially for TinyImageNet (almost $2\%$), we think the reasons are similar to those for open set recognition discussed above that complex datasets offer greater opportunities for contrastive learning to demonstrate its effects. 
Furthermore, with the introduction of GradMix, the models can focus better on saliency objects as shown in Figure \ref{fig-treser}.
We believe it can also explain why GradMix can bring improvements to robustness to corruptions and linear probing for SSL in \ref{exp-corruption} and \ref{subsec-exe-generalization}.
In conclusion, even though the introduction of SSL can lower the close set classification accuracy, GradMix can reduce this effect.

\subsection{Out of Distribution Detection} \label{subsec-exp-ood}

\paragraph{Settings} We validate our method on out-of-distribution detection (OOD) tasks. Following the settings for OOD in \cite{arpl, wang2024exploring}, we take CIFAR10 as in distribution set and CIFAR100 and SVHN as out of distribution sets. vanilla cross entropy, ARPL, ARPL-CS, and MEDAF are baselines. Evaluation metrics are TNR, AUROC, DTACC, and AUIN/AUOUT.
Details on complete settings are in Appendix \ref{app-details-ood}.

\paragraph{Results  } The results are given in table \ref{tab-ood-results}. GradMix can outperform the baselines, especially on SVHN, indicating that the model has learned many data-dependent features that overlap with the close OOD dataset CIFAR100.

\subsection{Generalization to Common Corruptions} \label{exp-corruption}

\noindent\textbf{Settings} 
We evaluate the robustness of the classifiers trained using our method on the data with common corruptions proposed in \cite{hendrycks2018benchmarking}, in which 15 types of corruptions are synthesized, namely Gaussian noise, shot noise, impulse noise, defocus blur, frosted glass blur, motion blur, zoom blur, snow, frost, fog, brightness, contrast, elastic, piexlate, and JPEG compression. Each of the corruption has five levels of severity and therefore the classifiers are tested with 75 repeats for all corruptions. We use accuracy drop, $D_{c, s}=A_{clean} - A_{c, s}$, as a metric to quantify the model's robustness to the corruption type $c$ of severity $s$, and $A_{clean}$ denotes the testing accuracy on data without corruptions. We record $D_{c, s}$ on the models trained with and without GradMix on full CIFAR10 and TinyImageNet datasets. Lower $D_{c, s}$ indicates higher robustness to the corruption (see Appendix \ref{app-details-corruptions} for more details). 

\noindent\textbf{Results  } 
The results are given in Figure \ref{fig-acc-corruption} and Table \ref{tab-acc-corruption}. In Figure \ref{fig-acc-corruption}, we average $A_{c, s}$ over the five severity levels for each corruption type, \ie, $\bar D_{c} = \frac{1}{5}\sum_{s} D_{c, s}$, and compare each $\bar D_{c}$ with and without GradMix. For most corruption types, $\bar D_{c}$s are smaller in the models with GradMix.
In Table \ref{tab-acc-corruption}, we average $A_{c, s}$ over the all corruption types for each severity level, \ie, $\bar D_{s} = \frac{1}{15}\sum_{c} D_{c, s}$. For almost all severity levels, $\bar D_{s}$ in models with GradMix are lower, indicating that GradMix classifiers are more robust to common corruptions.

\subsection{Model Generalization across Datasets} \label{subsec-exe-generalization}
\noindent\textbf{Settings} 
To validate the effectiveness of GradMix for model generalization, we apply it to self-supervised learning and test the downstream linear classification performances (linear probe). In order to increase the experimental diversity, the models are trained with SimCLR and MoCo v1 \cite{he2020momentum} as well as multiple network architectures of ResNet18, ResNet34, and ResNet50. 
Furthermore, in order to evaluate the effectiveness of GradMix for high-resolution data, we conduct the linear probe on the models trained on ImageNet100 \cite{rebuffi2017icarl}. We record top-1 and top-5 accuracy. Detailed settings are given in Appendix \ref{app-details-generalization}. 

\begin{figure}
    \includegraphics[width=\linewidth, height=0.3\textwidth]{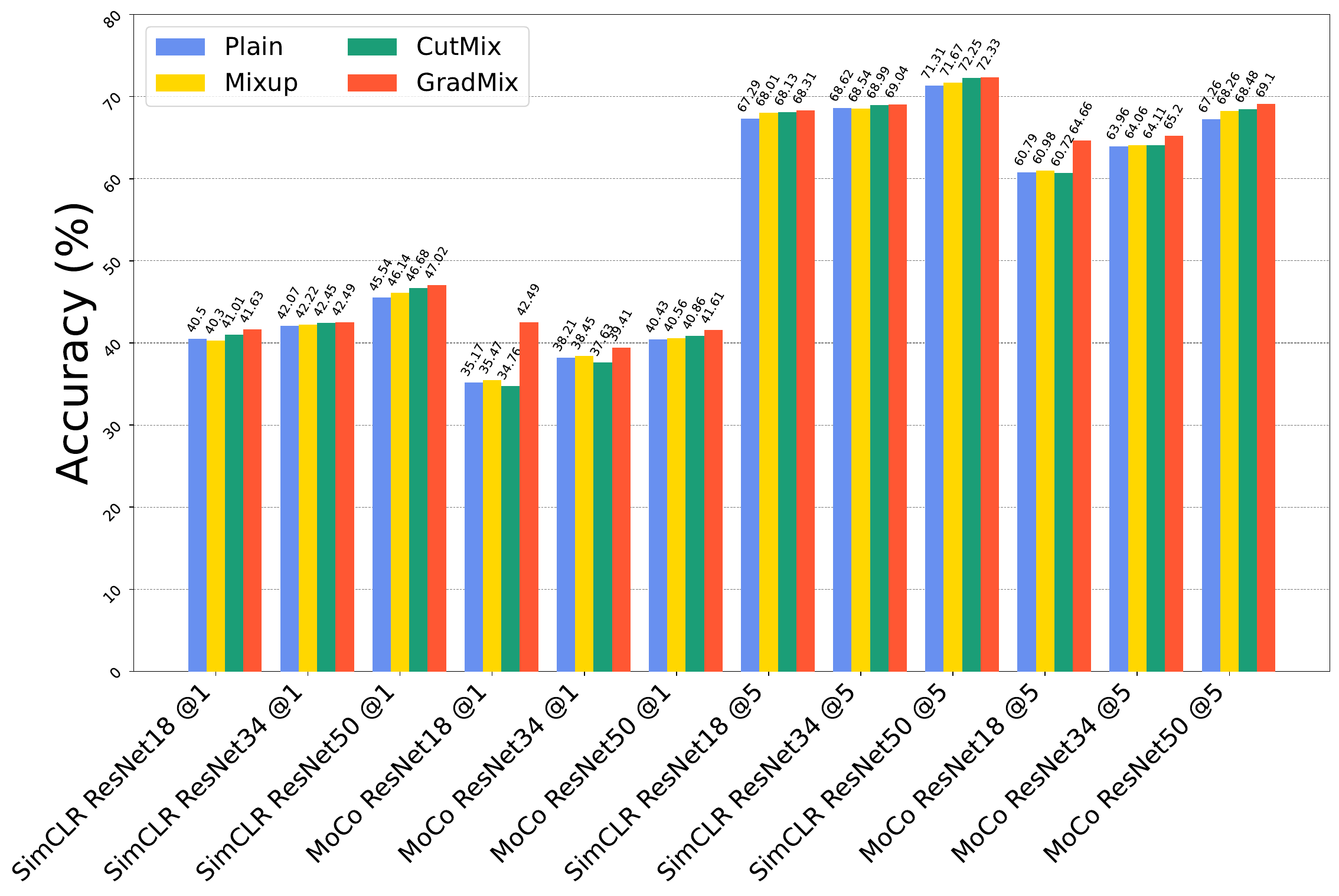}
  \caption{Top-1 and top-5 linear classification accuracy (in $\%$) on TinyImageNet with features learned using self-supervised learning with different augmentation methods. The results show that extra data augmentation can increase the downstream linear classification accuracy for SSL. GradMix can bring stronger improvements than other augmentation methods,  indicating its benefit for model generalization.}
  
  \label{fig-acc-ssl}
\end{figure}

\paragraph{Results  } The results are shown in Figure \ref{fig-acc-ssl} and Table \ref{tab-ssl-imaget100} for TinyImageNet and ImageNet100 datasets respectively. 
Both top-1 and top-5 accuracy increase with the application of extra augmentation methods and the improvements brought by GradMix are consistently higher. Qualitative visualizations of GradMix attribution maps from LayerCam on ImageNet100 are given in \cref{app-fig-imagenet100} in the Appendix \ref{app-imagenet100}.

\begin{table}[ht]
        \centering
        \fontsize{6pt}{6pt}\selectfont
        \begin{tabular}{ccccc}
        \toprule
         \multirow{2}{*}{Augmentations}        & \multicolumn{2}{c}{SimCLR} & \multicolumn{2}{c}{MoCo v1} \\
                                   & Acc@1 & Acc@5 & Acc@1 & Acc@5 \\
        \hline
           -            & 48.1 & 75.78 & 40.31 &  66.54 \\
        + Mixup         & 49.32 & 77.43  & 42.53 & 67.1 \\
        + CutMix        & 50.61 & 78.67 &  44.87 & 69.21 \\
        + GradMix       & \textbf{54.8} & \textbf{81.36}  & \textbf{47.75} & \textbf{72.37}\\
        \bottomrule
       \end{tabular}
       
\caption{Results from the linear probe on SSL models trained on ImageNet100 show that additional mixing data augmentations can lead to performance gains, with GradMix consistently achieving the best results.}
\label{tab-ssl-imaget100}
\end{table}

\section{Analysis \& Discussion}   \label{sec-analysis-discussion}
To further assess the effectiveness of GradMix in the above introduced supervised learning paradigm for OSR, we visualize the feature activations using LayerGAM. A selection of samples from CIFAR10 and TinyImageNet is shown in Figure \ref{fig-visualization}, with additional examples provided in Appendix \ref{app-more-visualizations}. It can be observed that the models trained with GradMix show larger activated areas, suggesting that GradMix enables the models to focus on a broader range of regions in the data, and to learne more diverse and informative features.

\begin{figure}[t]
\centering
    \includegraphics[trim={3.6cm 1cm 3.6cm 1cm},clip,width=0.16\linewidth]{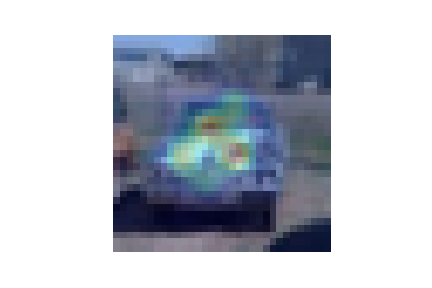}
    \includegraphics[trim={3.6cm 1cm 3.6cm 1cm},clip,width=0.16\linewidth]{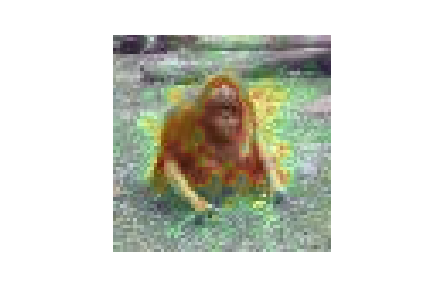}
    \includegraphics[trim={3.6cm 1cm 3.6cm 1cm},clip,width=0.16\linewidth]{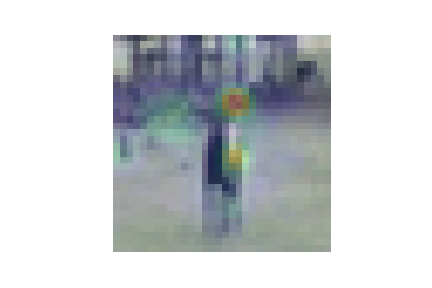}
    \includegraphics[trim={3.6cm 1cm 3.6cm 1cm},clip,width=0.16\linewidth]{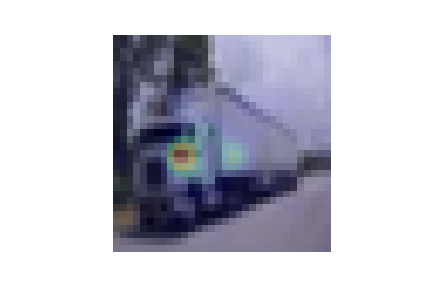}
    \\
    \includegraphics[trim={3.6cm 1cm 3.6cm 1cm},clip,width=0.16\linewidth]{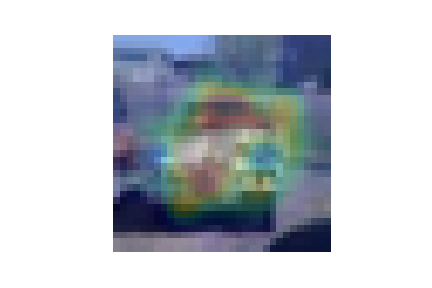}
    \includegraphics[trim={3.6cm 1cm 3.6cm 1cm},clip,width=0.16\linewidth]{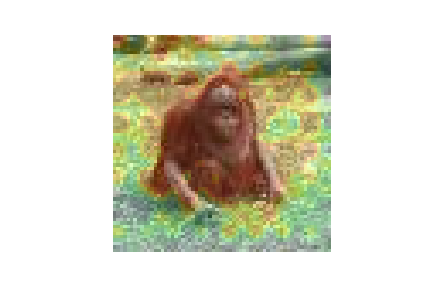}
    \includegraphics[trim={3.6cm 1cm 3.6cm 1cm},clip,width=0.16\linewidth]{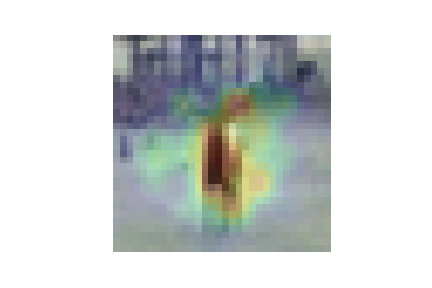}
    \includegraphics[trim={3.6cm 1cm 3.6cm 1cm},clip,width=0.16\linewidth]{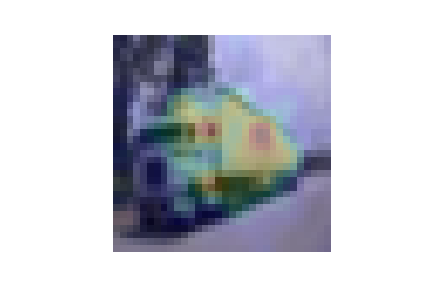}
    \caption{Attribution maps visualizations using LayerCAM: \textbf{First Row:} model trained without GradMix; \textbf{Second Row:} model trained with GradMix. The GradMix trained model has a focused activation that covers a larger part of the object. More attribution maps are given in \cref{app-fig-visualization} in \cref{app-more-visualizations} to confirm this qualitative observation.}
    \label{fig-visualization}
\end{figure}

\begin{figure}[ht]
    \centering
    \includegraphics[width=0.7\linewidth]{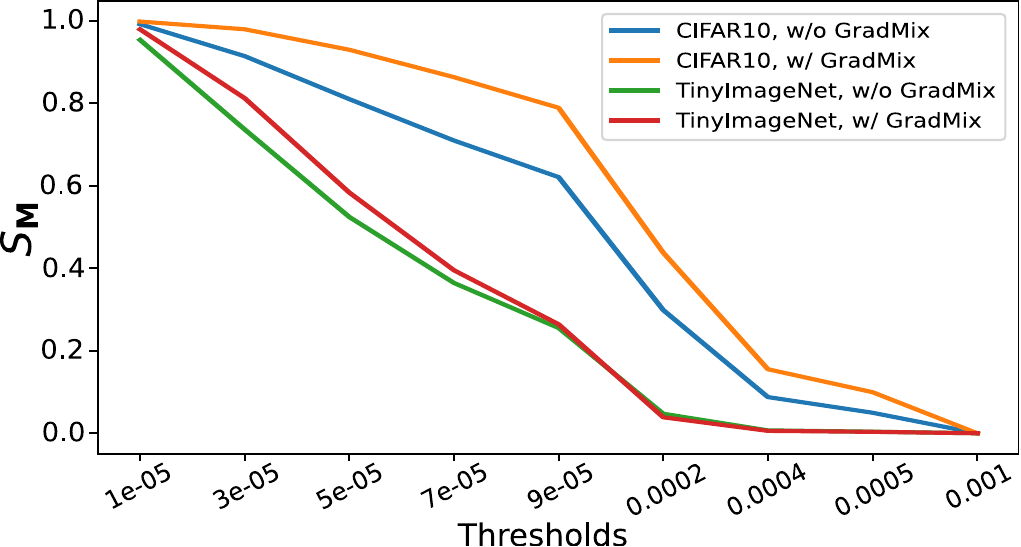}
    \caption{Change of $S_\mathbf{M}$ with $\tau$. The values of GradMix models are always higher, indicating broader activated areas.}
    \label{fig-activations}
\end{figure}

To quantitatively assess the activated areas in the data, we measure the number of higher-valued pixels in the attribution maps, denoted as $C_{\mathbf{M}}=\sum_{i,j} \mathbbm{1}_{\mathbf{M}_{i,j} > \tau}$, where $\tau$ is a predefined threshold. We vary $\tau$ from $10^{-5}$ to $10^{-3}$ and plot the fraction of $C_{\mathbf{M}}$ to the resolution of $\mathbf{M}$ (denoted as $S_\mathbf{M} = \frac{C_{\mathbf{M}}}{|\mathbf{M}|}$) in Figure \ref{fig-visualization}. $S_M$ with GradMix is always larger.
Consistent with the visualizations, GradMix leads the models to focus on a larger portion of the data, suggesting it helps capture a broader range of features.

\paragraph{Impact of Foundation Models}
With the popularity of foundation models (FMs), they have been repeatedly applied to OOD problems \cite{ming2022delving, li2024learning, miyai2024locoop, wang2023clipn}, which is similar to OSR. However, there are very few OSR solutions relying on FMs.  We believe one of the main reasons is that many novel classes are capsulated in the training sets of FMs, which violates the problem settings of OSR. Moreover, although FMs are trained using web-scale data, they can not reflect \emph{everything} in the real world. Most OOD solutions with FMs rely on vision-language models (VLMs) (see \ref{app-fm-ood} in Appendix), such as CLIP \cite{radford2021learning}, which are based on the premise that the multimodal data are completely aligned in latent space, which is, however, a question under research \cite{yarom2024you}. Furthermore, our linear probing results on ImageNet-100 in Table \ref{tab-ssl-imaget100} indicate that GradMix holds promise for improved model pre-training beyond OSR.

\section{Conclusion \& Outlook}  \label{sec-conclusion}
In this work, we proposed a novel approach for open set recognition, combining self-supervised learning with a gradient-based data augmentation method, guided by the idea of learning diverse features. Experimental results demonstrate that our approach surpasses most state-of-the-art methods in open set recognition, closed set classification, and out of distribution detection. Additionally, GradMix enhances model robustness against common corruptions and boosts downstream linear classification performance in self-supervised learning, further highlighting its effectiveness for improving model generalization. Our code will be published upon acceptance.

Recent works have reported that model sparsity can increase model robustness to adversarial attacks \cite{timpl2022understanding}. Besides the experiments in \ref{exp-corruption} on common data corruption, GradMix could be further evaluated for its effectiveness on adversarial attacks, which is beyond our current study.
Furthermore, as analyzed in \cite{chen2022sparsity}, model sparsification can allow new neural connections to grow and help the models to escape bad local minima, and hence reduce overfitting. It remains to explore if GradMix can achieve the same effects.

\bibliography{example_paper}

\begin{thebibliography}{53}
\providecommand{\natexlab}[1]{#1}
\providecommand{\url}[1]{\texttt{#1}}
\expandafter\ifx\csname urlstyle\endcsname\relax
  \providecommand{\doi}[1]{doi: #1}\else
  \providecommand{\doi}{doi: \begingroup \urlstyle{rm}\Url}\fi

\bibitem[Ansuini et~al.(2019)Ansuini, Laio, Macke, and Zoccolan]{ansuini2019intrinsic}
Ansuini, A., Laio, A., Macke, J.~H., and Zoccolan, D.
\newblock Intrinsic dimension of data representations in deep neural networks.
\newblock \emph{NeurIPS}, 32, 2019.

\bibitem[Bendale \& Boult(2016)Bendale and Boult]{bendale2016towards}
Bendale, A. and Boult, T.~E.
\newblock Towards open set deep networks.
\newblock In \emph{CVPR}, pp.\  1563--1572, 2016.

\bibitem[Cao et~al.(2021)Cao, Luo, and Klabjan]{cao2021open}
Cao, A., Luo, Y., and Klabjan, D.
\newblock Open-set recognition with gaussian mixture variational autoencoders.
\newblock In \emph{AAAI}, volume~35, pp.\  6877--6884, 2021.

\bibitem[Chen et~al.(2022{\natexlab{a}})Chen, Peng, Wang, and Tian]{arpl}
Chen, G., Peng, P., Wang, X., and Tian, Y.
\newblock Adversarial reciprocal points learning for open set recognition.
\newblock \emph{IEEE TPAMI}, 2022{\natexlab{a}}.

\bibitem[Chen et~al.(2022{\natexlab{b}})Chen, Fu, Narayan, Zhang, Song, Fatahalian, and R{\'e}]{chen2022perfectly}
Chen, M., Fu, D.~Y., Narayan, A., Zhang, M., Song, Z., Fatahalian, K., and R{\'e}, C.
\newblock Perfectly balanced: Improving transfer and robustness of supervised contrastive learning.
\newblock In \emph{International Conference on Machine Learning}, pp.\  3090--3122. PMLR, 2022{\natexlab{b}}.

\bibitem[Chen et~al.(2020)Chen, Kornblith, Norouzi, and Hinton]{chen2020simple}
Chen, T., Kornblith, S., Norouzi, M., and Hinton, G.
\newblock A simple framework for contrastive learning of visual representations.
\newblock In \emph{International Conference on Machine Learning}, pp.\  1597--1607. PMLR, 2020.

\bibitem[Chen et~al.(2022{\natexlab{c}})Chen, Zhang, Wang, Balachandra, Ma, Wang, and Wang]{chen2022sparsity}
Chen, T., Zhang, Z., Wang, P., Balachandra, S., Ma, H., Wang, Z., and Wang, Z.
\newblock Sparsity winning twice: Better robust generalization from more efficient training.
\newblock \emph{arXiv preprint arXiv:2202.09844}, 2022{\natexlab{c}}.

\bibitem[Deng et~al.(2009)Deng, Dong, Socher, Li, Li, and Fei-Fei]{deng2009imagenet}
Deng, J., Dong, W., Socher, R., Li, L.-J., Li, K., and Fei-Fei, L.
\newblock Imagenet: A large-scale hierarchical image database.
\newblock In \emph{CVPR}, pp.\  248--255. IEEE, 2009.

\bibitem[Deng(2012)]{deng2012mnist}
Deng, L.
\newblock The mnist database of handwritten digit images for machine learning research.
\newblock \emph{IEEE Sign. Process. Letters}, 29\penalty0 (6):\penalty0 141--142, 2012.

\bibitem[DeVries \& Taylor(2017)DeVries and Taylor]{devries2017improved}
DeVries, T. and Taylor, G.~W.
\newblock Improved regularization of convolutional neural networks with cutout.
\newblock \emph{arXiv preprint arXiv:1708.04552}, 2017.

\bibitem[Dhamija et~al.(2018)Dhamija, G{\"u}nther, and Boult]{dhamija2018reducing}
Dhamija, A.~R., G{\"u}nther, M., and Boult, T.
\newblock Reducing network agnostophobia.
\newblock \emph{NeurIPS}, 31, 2018.

\bibitem[Ge et~al.(2017)Ge, Demyanov, Chen, and Garnavi]{GOpenMax_Ge17}
Ge, Z., Demyanov, S., Chen, Z., and Garnavi, R.
\newblock Generative openmax for multi-class open set classification.
\newblock In \emph{BMVC}, London, UK, Sep. 2017.

\bibitem[Grill et~al.(2020)Grill, Strub, Altch{\'e}, Tallec, Richemond, Buchatskaya, Doersch, Avila~Pires, Guo, Gheshlaghi~Azar, et~al.]{grill2020bootstrap}
Grill, J.-B., Strub, F., Altch{\'e}, F., Tallec, C., Richemond, P., Buchatskaya, E., Doersch, C., Avila~Pires, B., Guo, Z., Gheshlaghi~Azar, M., et~al.
\newblock Bootstrap your own latent-a new approach to self-supervised learning.
\newblock \emph{NeurIPS}, 33:\penalty0 21271--21284, 2020.

\bibitem[Hassen \& Chan(2020)Hassen and Chan]{hassen2020learning}
Hassen, M. and Chan, P.~K.
\newblock Learning a neural-network-based representation for open set recognition.
\newblock In \emph{Proceedings of the 2020 SIAM International Conference on Data Mining}, pp.\  154--162. SIAM, 2020.

\bibitem[He et~al.(2016)He, Zhang, Ren, and Sun]{he2016deep}
He, K., Zhang, X., Ren, S., and Sun, J.
\newblock Deep residual learning for image recognition.
\newblock In \emph{CVPR}, pp.\  770--778, 2016.

\bibitem[He et~al.(2020)He, Fan, Wu, Xie, and Girshick]{he2020momentum}
He, K., Fan, H., Wu, Y., Xie, S., and Girshick, R.
\newblock Momentum contrast for unsupervised visual representation learning.
\newblock In \emph{CVPR}, pp.\  9729--9738, 2020.

\bibitem[Hendrycks \& Dietterich(2018)Hendrycks and Dietterich]{hendrycks2018benchmarking}
Hendrycks, D. and Dietterich, T.
\newblock Benchmarking neural network robustness to common corruptions and perturbations.
\newblock In \emph{ICLR}, 2018.

\bibitem[Jiang et~al.(2021)Jiang, Zhang, Hou, Cheng, and Wei]{jiang2021layercam}
Jiang, P.-T., Zhang, C.-B., Hou, Q., Cheng, M.-M., and Wei, Y.
\newblock Layercam: Exploring hierarchical class activation maps for localization.
\newblock \emph{IEEE TIP}, 30:\penalty0 5875--5888, 2021.

\bibitem[Jing et~al.(2021)Jing, Vincent, LeCun, and Tian]{jing2021understanding}
Jing, L., Vincent, P., LeCun, Y., and Tian, Y.
\newblock Understanding dimensional collapse in contrastive self-supervised learning.
\newblock \emph{arXiv preprint arXiv:2110.09348}, 2021.

\bibitem[Khosla et~al.(2020)Khosla, Teterwak, Wang, Sarna, Tian, Isola, Maschinot, Liu, and Krishnan]{khosla2020supervised}
Khosla, P., Teterwak, P., Wang, C., Sarna, A., Tian, Y., Isola, P., Maschinot, A., Liu, C., and Krishnan, D.
\newblock Supervised contrastive learning.
\newblock \emph{NeurIPS}, 33:\penalty0 18661--18673, 2020.

\bibitem[Krizhevsky et~al.(2009{\natexlab{a}})Krizhevsky, Hinton, et~al.]{cifar10}
Krizhevsky, A., Hinton, G., et~al.
\newblock Learning multiple layers of features from tiny images, 2009{\natexlab{a}}.

\bibitem[Krizhevsky et~al.(2009{\natexlab{b}})Krizhevsky, Hinton, et~al.]{cifar100}
Krizhevsky, A., Hinton, G., et~al.
\newblock Learning multiple layers of features from tiny images, 2009{\natexlab{b}}.

\bibitem[Lewy \& Ma{\'n}dziuk(2023)Lewy and Ma{\'n}dziuk]{lewy2023overview}
Lewy, D. and Ma{\'n}dziuk, J.
\newblock An overview of mixing augmentation methods and augmentation strategies.
\newblock \emph{Artificial Intelligence Review}, 56\penalty0 (3):\penalty0 2111--2169, 2023.

\bibitem[Li et~al.(2024)Li, Pang, Bai, Miao, and Zheng]{li2024learning}
Li, T., Pang, G., Bai, X., Miao, W., and Zheng, J.
\newblock Learning transferable negative prompts for out-of-distribution detection.
\newblock In \emph{CVPR}, pp.\  17584--17594, 2024.

\bibitem[Mahdavi \& Carvalho(2021)Mahdavi and Carvalho]{mahdavi2021survey}
Mahdavi, A. and Carvalho, M.
\newblock A survey on open set recognition.
\newblock In \emph{2021 IEEE Fourth International Conference on Artificial Intelligence and Knowledge Engineering (AIKE)}, pp.\  37--44. IEEE, 2021.

\bibitem[Miller et~al.(2021)Miller, Sunderhauf, Milford, and Dayoub]{miller2021class}
Miller, D., Sunderhauf, N., Milford, M., and Dayoub, F.
\newblock Class anchor clustering: A loss for distance-based open set recognition.
\newblock In \emph{Proceedings of the IEEE/CVF Winter Conference on Applications of Computer Vision}, pp.\  3570--3578, 2021.

\bibitem[Ming et~al.(2022)Ming, Cai, Gu, Sun, Li, and Li]{ming2022delving}
Ming, Y., Cai, Z., Gu, J., Sun, Y., Li, W., and Li, Y.
\newblock Delving into out-of-distribution detection with vision-language representations.
\newblock \emph{NeurIPS}, 35:\penalty0 35087--35102, 2022.

\bibitem[Miyai et~al.(2024)Miyai, Yu, Irie, and Aizawa]{miyai2024locoop}
Miyai, A., Yu, Q., Irie, G., and Aizawa, K.
\newblock Locoop: Few-shot out-of-distribution detection via prompt learning.
\newblock \emph{NeurIPS}, 36, 2024.

\bibitem[Neal et~al.(2018)Neal, Olson, Fern, Wong, and Li]{neal2018open}
Neal, L., Olson, M., Fern, X., Wong, W.-K., and Li, F.
\newblock Open set learning with counterfactual images.
\newblock In \emph{ECCV}, pp.\  613--628, 2018.

\bibitem[Netzer et~al.(2011)Netzer, Wang, Coates, Bissacco, Wu, and Y.Ng]{Netzer2011SVHN}
Netzer, Y., Wang, T., Coates, A., Bissacco, A., Wu, B., and Y.Ng, A.
\newblock Street view hause number dataset.
\newblock \url{http://ufldl.stanford.edu/housenumbers}, 2011.

\bibitem[Oord et~al.(2018)Oord, Li, and Vinyals]{oord2018representation}
Oord, A. v.~d., Li, Y., and Vinyals, O.
\newblock Representation learning with contrastive predictive coding.
\newblock \emph{arXiv preprint arXiv:1807.03748}, 2018.

\bibitem[Oza \& Patel(2019)Oza and Patel]{oza2019c2ae}
Oza, P. and Patel, V.~M.
\newblock C2ae: Class conditioned auto-encoder for open-set recognition.
\newblock In \emph{CVPR}, pp.\  2307--2316, 2019.

\bibitem[Perera et~al.(2020)Perera, Morariu, Jain, Manjunatha, Wigington, Ordonez, and Patel]{perera2020generative}
Perera, P., Morariu, V.~I., Jain, R., Manjunatha, V., Wigington, C., Ordonez, V., and Patel, V.~M.
\newblock Generative-discriminative feature representations for open-set recognition.
\newblock In \emph{CVPR}, pp.\  11814--11823, 2020.

\bibitem[Radford et~al.(2021)Radford, Kim, Hallacy, Ramesh, Goh, Agarwal, Sastry, Askell, Mishkin, Clark, et~al.]{radford2021learning}
Radford, A., Kim, J.~W., Hallacy, C., Ramesh, A., Goh, G., Agarwal, S., Sastry, G., Askell, A., Mishkin, P., Clark, J., et~al.
\newblock Learning transferable visual models from natural language supervision.
\newblock In \emph{ICLR}, pp.\  8748--8763. PMLR, 2021.

\bibitem[Rebuffi et~al.(2017)Rebuffi, Kolesnikov, Sperl, and Lampert]{rebuffi2017icarl}
Rebuffi, S.-A., Kolesnikov, A., Sperl, G., and Lampert, C.~H.
\newblock icarl: Incremental classifier and representation learning.
\newblock In \emph{CVPR}, pp.\  2001--2010, 2017.

\bibitem[Russakovsky et~al.(2015)Russakovsky, Deng, Su, Krause, Satheesh, Ma, Huang, Karpathy, Khosla, Bernstein, et~al.]{russakovsky2015imagenet}
Russakovsky, O., Deng, J., Su, H., Krause, J., Satheesh, S., Ma, S., Huang, Z., Karpathy, A., Khosla, A., Bernstein, M., et~al.
\newblock Imagenet large scale visual recognition challenge.
\newblock \emph{IJCV}, 115:\penalty0 211--252, 2015.

\bibitem[Scheirer et~al.(2012)Scheirer, de~Rezende~Rocha, Sapkota, and Boult]{scheirer2012toward}
Scheirer, W.~J., de~Rezende~Rocha, A., Sapkota, A., and Boult, T.~E.
\newblock Toward open set recognition.
\newblock \emph{IEEE TPAMI}, 35\penalty0 (7):\penalty0 1757--1772, 2012.

\bibitem[Selvaraju et~al.(2017)Selvaraju, Cogswell, Das, Vedantam, Parikh, and Batra]{selvaraju2017grad}
Selvaraju, R.~R., Cogswell, M., Das, A., Vedantam, R., Parikh, D., and Batra, D.
\newblock Grad-cam: Visual explanations from deep networks via gradient-based localization.
\newblock In \emph{ICCV}, pp.\  618--626, 2017.

\bibitem[Shu \& Lampos(2024)Shu and Lampos]{shu2024unsupervised}
Shu, Y. and Lampos, V.
\newblock Unsupervised hard negative augmentation for contrastive learning.
\newblock \emph{arXiv preprint arXiv:2401.02594}, 2024.

\bibitem[Timpl et~al.(2022)Timpl, Entezari, Sedghi, Neyshabur, and Saukh]{timpl2022understanding}
Timpl, L., Entezari, R., Sedghi, H., Neyshabur, B., and Saukh, O.
\newblock Understanding the effect of sparsity on neural networks robustness.
\newblock \emph{arXiv preprint arXiv:2206.10915}, 2022.

\bibitem[Vaze et~al.(2022)Vaze, Han, Vedaldi, and Zisserman]{vaze2022openset}
Vaze, S., Han, K., Vedaldi, A., and Zisserman, A.
\newblock Open-set recognition: a good closed-set classifier is all you need?
\newblock In \emph{ICLR}, 2022.

\bibitem[Walawalkar et~al.(2020)Walawalkar, Shen, Liu, and Savvides]{walawalkar2020attentive}
Walawalkar, D., Shen, Z., Liu, Z., and Savvides, M.
\newblock Attentive cutmix: An enhanced data augmentation approach for deep learning based image classification.
\newblock \emph{arXiv preprint arXiv:2003.13048}, 2020.

\bibitem[Wang et~al.(2023)Wang, Li, Yao, and Li]{wang2023clipn}
Wang, H., Li, Y., Yao, H., and Li, X.
\newblock Clipn for zero-shot ood detection: Teaching clip to say no.
\newblock In \emph{ICCV}, pp.\  1802--1812, 2023.

\bibitem[Wang et~al.(2024)Wang, Mu, Zhu, and Hu]{wang2024exploring}
Wang, Y., Mu, J., Zhu, P., and Hu, Q.
\newblock Exploring diverse representations for open set recognition.
\newblock In \emph{AAAI}, volume~38, pp.\  5731--5739, 2024.

\bibitem[Wang et~al.(2022)Wang, Xu, Yang, He, Cao, and Huang]{wang2022openauc}
Wang, Z., Xu, Q., Yang, Z., He, Y., Cao, X., and Huang, Q.
\newblock Openauc: Towards auc-oriented open-set recognition.
\newblock \emph{NeurIPS}, 35:\penalty0 25033--25045, 2022.

\bibitem[Xu et~al.(2023)Xu, Shen, and Zhao]{xu2023contrastive}
Xu, B., Shen, F., and Zhao, J.
\newblock Contrastive open set recognition.
\newblock In \emph{AAAI}, volume~37, pp.\  10546--10556, 2023.

\bibitem[Xue et~al.(2023)Xue, Joshi, Gan, Chen, and Mirzasoleiman]{xue2023features}
Xue, Y., Joshi, S., Gan, E., Chen, P.-Y., and Mirzasoleiman, B.
\newblock Which features are learnt by contrastive learning? on the role of simplicity bias in class collapse and feature suppression.
\newblock In \emph{International Conference on Machine Learning}, pp.\  38938--38970. PMLR, 2023.

\bibitem[Yarom et~al.(2024)Yarom, Bitton, Changpinyo, Aharoni, Herzig, Lang, Ofek, and Szpektor]{yarom2024you}
Yarom, M., Bitton, Y., Changpinyo, S., Aharoni, R., Herzig, J., Lang, O., Ofek, E., and Szpektor, I.
\newblock What you see is what you read? improving text-image alignment evaluation.
\newblock \emph{NeurIPS}, 36, 2024.

\bibitem[Yoshihashi et~al.(2019)Yoshihashi, Shao, Kawakami, You, Iida, and Naemura]{yoshihashi2019classification}
Yoshihashi, R., Shao, W., Kawakami, R., You, S., Iida, M., and Naemura, T.
\newblock Classification-reconstruction learning for open-set recognition.
\newblock In \emph{CVPR}, pp.\  4016--4025, 2019.

\bibitem[Yu et~al.(2017)Yu, Qu, Li, and Guo]{asg_Yu17}
Yu, Y., Qu, W.-Y., Li, N., and Guo, Z.
\newblock Open category classification by adversarial sample generation.
\newblock In \emph{IJCAI}, pp.\  3357--3363, Melbourne, Australia, Aug. 2017.

\bibitem[Yun et~al.(2019)Yun, Han, Oh, Chun, Choe, and Yoo]{yun2019cutmix}
Yun, S., Han, D., Oh, S.~J., Chun, S., Choe, J., and Yoo, Y.
\newblock Cutmix: Regularization strategy to train strong classifiers with localizable features.
\newblock In \emph{CVPR}, pp.\  6023--6032, 2019.

\bibitem[Zhang et~al.(2017)Zhang, Cisse, Dauphin, and Lopez-Paz]{zhang2017mixup}
Zhang, H., Cisse, M., Dauphin, Y.~N., and Lopez-Paz, D.
\newblock mixup: Beyond empirical risk minimization.
\newblock \emph{arXiv preprint arXiv:1710.09412}, 2017.

\bibitem[Zhou et~al.(2021)Zhou, Ye, and Zhan]{zhou2021learning}
Zhou, D.-W., Ye, H.-J., and Zhan, D.-C.
\newblock Learning placeholders for open-set recognition.
\newblock In \emph{CVPR}, pp.\  4401--4410, 2021.

\end{thebibliography}
\bibliographystyle{icml2025}

\newpage~
\newpage~
\appendix

\section{Details of the Experiments in Section \ref{sec-introduction}} \label{app-introduction-exp}

For a more thorough investigation of the effectiveness of applying self-supervised learning (SSL) within supervised paradigms, we trained four sets of models on the CIFAR10 and TinyImageNet datasets. These models were optimized with the following learning objectives: supervised contrastive learning (SupCon), supervised contrastive learning combined with self-supervised contrastive learning (SupCon+SSL), cross-entropy classification (CE), and cross-entropy combined with self-supervised contrastive learning (CE+SSL).
SupCon + SSL is introduced in \ref{sec-diverse-feature-learning} (see Equ. \eqref{equ-cl}).
The SupCon+SSL objective is introduced in Section \ref{sec-diverse-feature-learning} (see Equation \eqref{equ-cl}). For CE+SSL, SSL is applied to the final layer of the backbone encoder. Similar to SupCon, the final learning objective is a convex combination of the two losses, i.e., $L_{ce+ssl} = \gamma * L_{ce} +  \lambda *  L_{ssl}$. where the backbone encoder is directly followed by the classification head. For both CE+SSL and SupCon+SSL, we set $\gamma=1, \lambda=1$. All models use ResNet18 as the backbone, with a feature dimension of 128 for contrastive learning. To ensure fairness, all models were trained for 600 epochs using default augmentations for contrastive learning as specified in their respective original papers. As with the OSR experiments described in Section \ref{subsec-exp-osr}, the final results are averaged over five trials with different open and closed set splits.

Additionally, we employ multiple detection methods to identify open set samples, aiming to enhance the comprehensiveness of our experiments. For $A_{enp}$ and $A_{msp}$, the model's output logits for each test sample are first converted into softmax probabilities. The Shannon entropy of these probabilities and the maximum probability are then used as detection scores, respectively.
For $A_{nno}$, we adopt a modified approach based on \cite{scheirer2012toward}, eliminating the metric learning phase to directly evaluate the features learned by the neural networks. In this method, the Mahalanobis distance between the test sample's feature and its closest class mean serves as the detection score.

\section{A Comparison between Attentive CutMix and \emph{GradMix}} \label{app-compare}

\begin{figure}
  \includegraphics[width=\linewidth]{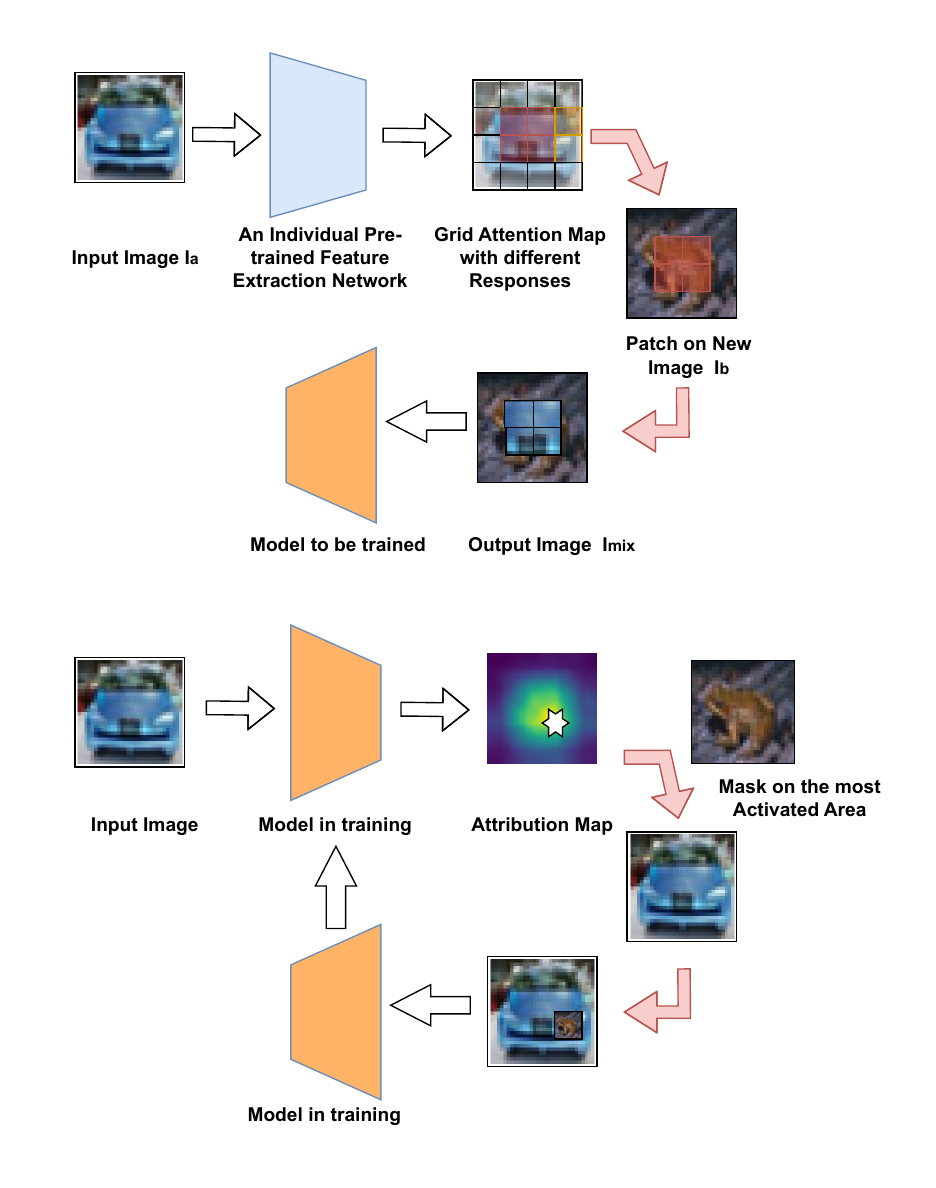}
 \caption{A graphical comparison between Attentive Mixup (top) and \emph{GradMix} (bottom) is illustrated. Attentive Mixup identifies high-response grids in input images by utilizing feature maps from a pre-trained model. These identified grids are then patched on the same spatial positions in a new image. And the new image with patches is the input for training the model. In contrast, \emph{GradMix} masks the most activated regions in the original image using attribution maps derived from the training model itself, replacing them with patches from a randomly selected image within the same mini-batch. The masked original image is then used for training.}
 \label{fig-app-compare}
 \end{figure}

As mentioned in \ref{subsec-method-gradmix}, \emph{GradMix} and Attentive CutMix are based on different principles. 
A graphical comparison between Attentive Mixup and GradMix is in Figure \ref{fig-app-compare}. 
Attentive Mix is specifically designed for supervised learning, the most attentive regions in image $\mathbf{I}_a$ with label $l_a$ discovered by the pre-trained models are patched on a new image $\mathbf{I}_b$ with label $l_b$. The mixed image $\mathbf{I}_{ab}$ is used for model training. It can guide the models to concentrate on these most attentive regions in the original images. Same as vanilla CutMix, the label for the $\mathbf{I}_{ab}$, $l_{ab}$, is the weighted sum of $l_a$ and $l_b$, \ie, 

\begin{equation}
    l_{ab} = \lambda l_a + (1-\lambda) l_b
\end{equation}

in which $\lambda$ stands for portion of $\mathbf{I}_a$ in $\mathbf{I}_{ab}$.

GradMix aims to mask out the highly activated (learned) regions in the data, encouraging the models to focus on broader image areas. Therefore, directly applying Attentive Mix to \emph{GradMix} or a self-supervised learning framework for comparison is not meaningful. In such a setting, the two views of the inputs should be $\mathbf{I}_{b}$ and $\mathbf{I}_{ab}$, which results in no distinction from vanilla CutMix. In other words, the attentive patched grids do not contribute effectively in this scenario.
Moreover, the pre-trained models used in Attentive CutMix are trained on ImageNet-1K, which contains semantic classes present in CIFAR and TinyImageNet. This introduces an unfair advantage when comparing it with other methods that do not leverage such extensive prior knowledge.

\section{More on the Detection Method} \label{app-detection-method}

Given a training set $\mathbi{X}_{\textit{train}}$, where $\mathbi{X}_{\textit{train}} = {\mathbi{X}_1,...,\mathbi{X}_C}$, with $C$ classes, and $\mathbi{Z}_\textit{train} = {\mathbi{Z}_1,...,\mathbi{Z}_C}$ denotes their corresponding feature representations, we compute the classification scores for the test set $\mathbi{X}_\textit{test}$ following algorithm \ref{alg-knn}.

For one testing sample $\mathbf{x}_i$ and its deep representation $\mathbf{z}_i$, we compute its similarity with each element of $\mathbi{Z}_c$ in $\mathbi{Z}_\textit{train}$, and select the top k largest similarities of each class and compute their summation to get $d_{i}^{c}$. 
The classification score $s_i$ for $\mathbf{x}_i$ is then the normalized $d_{i}^{c}$ among all in set classes.

\begin{algorithm}
\caption{Open Set Recognition Framework}
\label{alg-knn}
\begin{algorithmic}[1]
\STATE \textbf{Input}: Feature encoder $\mathbb{E}$, $\mathbi{X}_{train}$, $\mathbi{X}_{test}$, and hyper-parameter $k$
\STATE \textbf{Output}: Set of classification scores, $G_\mathit{test}$, and predicted labels, $\hat{Y}_\mathit{test}$, for each $\mathbf{x}_{i}$ in $\mathbi{X}_{test}$.
\STATE \textbf{Initialize}: $\hat{Y}_\mathit{test}=\emptyset$
\newline

\FOR{$\mathbf{x}_{i}$ in $ \mathbi{X}_{test}$}
     \STATE $\mathbf{z}_{i} = \mathbb{E}(\mathbf{x}_{i})$
     \STATE $G_{i}=\emptyset$
     \FOR{$\mathbf{Z}_{c}$ in $\mathbi{Z}_{train}$}
         \STATE Compute similarities between $\mathbf{z}_{i}$ and each element in $\mathbi{Z}_{c}$, $S_c = \mathbf{sim}(\mathbf{z}_\mathit{test}, \mathbf{Z_{c}})$
         \STATE Select the top $k$ values in $S_c$, $S_c^{k}=\mathbf{max}(S_c, k)$
         \STATE $d_{i}^{c} = \mathbf{sum}(S_c^{k})$
         \STATE $G_{i} = G_\mathit{i} \cup \{ d_{i}^{c} \} $
     \ENDFOR
     
     \STATE $\hat{y}_{i} = \mathbf{argmax}(G_{i})$
     \STATE $ s_{i} = \mathbf{max}_c \frac{d_{i}^{c}}{\sum_{d_{i}^{c} \in G_{i}} d_{i}^{c}}$
     \STATE $\hat{Y}_\mathit{test} = \hat{Y}_\mathit{test} \cup \{\hat{y}_{i} \}$
     
\ENDFOR
\end{algorithmic}  
\end{algorithm}

\section{Details on OSR experiments}  \label{app-osr-exp}

\subsection{Evaluation Protocols}

Following section \ref{subsec-exp-osr}, a description of the six splitting protocols for OSR is in Table \ref{tab-datasets-split}. Openness quantifies the degree to which a task includes unknown or novel classes during inference. It is a formal measure of the challenge posed by the presence of classes that were not seen during training. Assume that the number of known and unknown classes are denoted using $k$ and $u$ respectively, openness ($O$) is defined mathematically as:

\begin{equation}
    O = \sqrt{\frac{2 * k}{ k + u}}
\end{equation}

\begin{table}[h]
    \centering
    \fontsize{7pt}{7pt}\selectfont
    \begin{tabular}{cccc}
    \toprule
        Protocols & \makecell{Known \\ ($\sharp$ Classes / Source)} & \makecell{Unknown \\ ($\sharp$  Classes / Source)} &  \makecell{Openness} \\
        \hline
        MNIST         &   6 / MNIST    &   4 / MNIST     &  $22.54 \%$         \\
        SVHN          &   6 / SVHN     &   4 / SVHN      &  $22.54 \%$      \\
        CIFAR10       &   6 / CIFAR10   &  4 / CIFAR10   &  $22.54 \%$         \\
        CIFAR+10      &   4 / CIFAR10    &  10 / CIFAR100   &  $46.55 \%$          \\
        CIFAR+50      &   4 / CIFAR10   &     50 / CIFAR100   &  $72.784 \%$       \\
        TinyImagenet   & 20 / TinyImagenet     &   180 / TinyImagenet  &  $68.37 \%$     \\
    \bottomrule
    \end{tabular}
    \caption{Datasets splitting protocols for known and unknown classes applied for evaluating OSR performance in Section \ref{subsec-exp-osr} in the main text.}
    \label{tab-datasets-split}
\end{table}

\subsection{Hyper-parameters}   \label{osr-exp-hyperparameter}
For experiments in \ref{subsec-exp-osr} in the main text for open set recognition, we use ResNet18 as the backbone encoders and the output feature dimension is 128. Except for MNIST and SVHN, we aggregate layer $\textsf{conv3\_2}$, $\textsf{conv4\_2}$, $\textsf{conv5\_2}$ in ResNet18 to compute the attribution maps for GradMix.
We list the final applied hyper-parameters in Table \ref{tab-hyperparameter-osr}, including $\gamma$ and $\lambda$ in the learning objective, $k$ in the detection method, the number of training epochs, and batch size (BS). A cosine annealing scheduler is applied to adjust the learning rate with the maximum and minimum learning rate of $10^{-3}$ and $5.12*10^{-5}$ respectively. Adam optimizer is utilized to train the models.

\begin{table}[h]
    \centering
    \fontsize{7pt}{7pt}\selectfont
    \begin{tabular}{cccccc}
    \toprule
        Protocols & $\gamma$ & $\lambda$ & k &  Epochs  & BS  \\
        \hline
        MNIST         &   1.0 &   1.0    &  3    &  20 &  256 \\
        SVHN          &   1.0 &   1.0  &  3 &  100   &  256 \\
        CIFAR10       &   1.0 &   1.2 &   3 &  600  & 256 \\
        CIFAR+10      &   1.0 &   1.2  &  3  & 600  & 256 \\
        CIFAR+50      &   1.0 &   1.2 &    3  & 600  & 256 \\
        TinyImagenet   &  1.0 &   1.0  &   3  & 600  & 256 \\
    \bottomrule
    \end{tabular}
    \caption{Settings of hyper-parameters in \ref{subsec-exp-osr} of the main text.}
    \label{tab-hyperparameter-osr}
\end{table}

The CIFAR10 protocol settings are applied to models trained on CIFAR10 and CIFAR100 for closed-set classification in \ref{subsec-close-set} in the main text. Meanwhile, the TinyImageNet settings are used for models on TinyImageNet in the same section.

\subsection{Implementation Details of Attentive CutMix} \label{attentive-details-osr}
We leverage the open source code \footnote{\url{https://github.com/xden2331/attentive_cutmix/tree/main}} to evaluate Attentive Mixup in open set recognition. The pre-trained model used for computing the attention maps is ResNet18 trained with ImageNet-1k provided by \emph{PyTorch}. The rest settings follow the original paper \cite{walawalkar2020attentive}.

\section{Experimental Details for \ref{subsec-exp-ood}}   \label{app-details-ood}

\noindent\textbf{TNR} stands for the true negative rate when the true positive rate (TPR) is $95\%$. Let TP, TN, FP, FN represent true positive, true negative, false positive and false negative respectively (same for the following text), $\mathrm{TNR} = \mathrm{TN} / (\mathrm{TP}+\mathrm{TN})$ when $\mathrm{TPR}=\mathrm{TP} / (\mathrm{FP} + \mathrm{FN})$ is $95\%$.

\noindent\textbf{DTACC} is defined as the maximum classification accuracy that can be achieved by setting all possible thresholds, namely $\mathrm{DTACC}= (\mathrm{TP} + \mathrm{TN}) / ( \mathrm{TP} + \mathrm{TN} + \mathrm{FP} + \mathrm{FN})$

\noindent\textbf{AUIN or AUOUT} are the area under the precision-recall curve (AUPR) when in- or out-of-distribution samples are specified as positive respectively.

The experimental settings are similar to these in OSR experiments: ResNet18 acts as the backbone encoder with the output dimension of 128. Layer aggregation is applied for GradMix with layers $\textsf{conv3\_2}$, $\textsf{conv4\_2}$, $\textsf{conv5\_2}$ in ResNet18. Batch size is 256 and the model is trained with 600 epochs.

\section{Experimental Details for \ref{exp-corruption}}   \label{app-details-corruptions}

We train models on full clean CIFAR10 and TinyImageNet datasets and test their robustness to data corruption. The baseline models without GradMix are with the learning objective of combining supervised and self-supervised contrastive learning as introduced in \ref{sec-diverse-feature-learning}. For models with and without GradMix, $\lambda$ is set to 1.0 and 1.2 for CIFAR10 and TinyImageNet datasets respectively, and $\gamma=1$ for both. Batch sizes are 256.

All models are based on the ResNet18 backbone with the output feature dimension of 128 and trained with 600 epochs.  We use the detection method in \ref{subsec-method-osr-framework} during inference and $k$ is set to 3. Settings on learning rate and optimizers are identical as for OSR experiments in \ref{subsec-exp-osr} in the main text and \ref{app-osr-exp} in this supplementary material.

\section{Experimental Details for \ref{subsec-exe-generalization}}   \label{app-details-generalization}

SimCLR and MoCo v1 models are trained in \ref{subsec-exe-generalization} and linear probing is applied afterwards to evaluate their performance in downstream tasks. The models are trained on full TinyImageNet and ImageNet100 datasets. ImageNet100 is a subset of the ImageNet-1k dataset and contains 100 randomly selected classes.
ResNet18, ResNet34, and ResNet50 are the backbone architectures for TinyImageNet models, and ImageNet100 models rely on ResNet18. Table \ref{tab-epochs-ssl} lists the batch sizes and the number of epochs trained for each model (the settings are identical for models without and with different data augmentations, \ie, mixup, cutmix, and GradMix).

\begin{table}[h]
    \centering
    \fontsize{8pt}{8pt}\selectfont
    \begin{tabular}{ccccc}
    \toprule
        Datasets & Methods & Architectures & Epochs & BS \\
        \hline
        \multirow{6}{*}{TinyImageNet}         &   \multirow{3}{*}{SimCLR}    &  ResNet18    & 200 & 256  \\
                                              &     &  ResNet34 &  200 & 256 \\
                                              &     &  ResNet50 &  200 & 256 \\
        \addlinespace[1ex]
                                              &   \multirow{3}{*}{MoCo}    &  ResNet18    & 200 & 256 \\
                                              &     &  ResNet34 & 200 & 256 \\
                                              &     &  ResNet50 & 200 & 256 \\
        \addlinespace[3ex]
        \multirow{2}{*}{ImageNet100}         &   SimCLR    &  ResNet18    & 100 & 256 \\
                                              &   MoCo     &  ResNet18    & 100 & 256 \\
    \bottomrule
    \end{tabular}
    \caption{Number of training epochs for models in \ref{subsec-exe-generalization}. The settings are the same for models without and with different data augmentations, \ie, mixup, cutmix, and GradMix.}
    \label{tab-epochs-ssl}
\end{table}

For both MoCo and SimCLR models, the dimension of the output representations is 128. 
For MoCo models, the queue size and momentum for updating the key encoder are 4096 and 0.999 respectively. 
Extra data augmentation is applied as introduced in \eqref{equ-loss-all} excluding the SupCon part and the balancing hyper-parameters, \ie, $L= L_{\textit{ssl}} + \gamma^2 \cdot L_{\textit{ssl}}^{\textit{aug}}$. $L_{\textit{ssl}}$ denotes the standard loss function for SimCLR or MoCo.

\section{Attribution Maps for Section \ref{subsec-exe-generalization} on ImageNet100} \label{app-imagenet100}

\begin{figure}[ht]
\centering
    \includegraphics[trim={3.6cm 1cm 3.6cm 1cm},clip,width=0.3\linewidth]{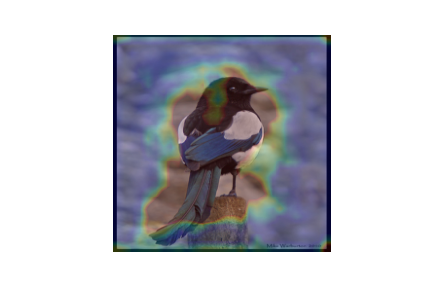}
    \includegraphics[trim={3.6cm 1cm 3.6cm 1cm},clip,width=0.3\linewidth]{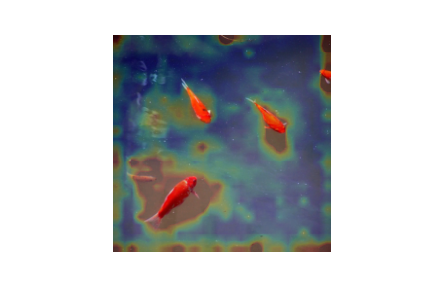}
    \includegraphics[trim={3.6cm 1cm 3.6cm 1cm},clip,width=0.3\linewidth]{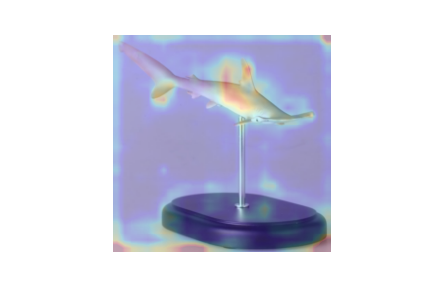}
    \\
    \includegraphics[trim={3.6cm 1cm 3.6cm 1cm},clip,width=0.3\linewidth]{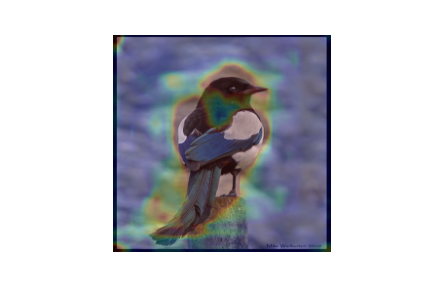}
    \includegraphics[trim={3.6cm 1cm 3.6cm 1cm},clip,width=0.3\linewidth]{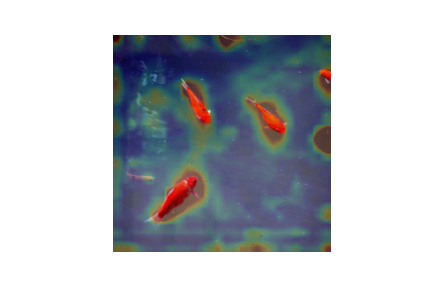}
    \includegraphics[trim={3.6cm 1cm 3.6cm 1cm},clip,width=0.3\linewidth]{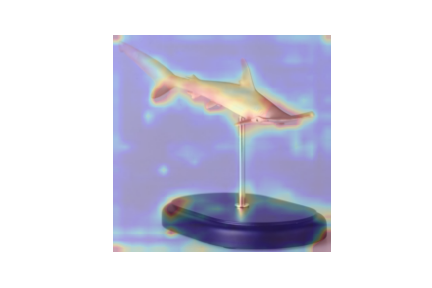}
    \\
    \includegraphics[trim={3.6cm 1cm 3.6cm 1cm},clip,width=0.3\linewidth]{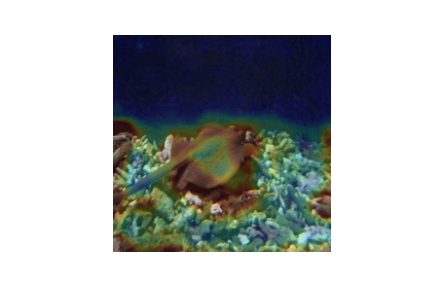}
    \includegraphics[trim={3.6cm 1cm 3.6cm 1cm},clip,width=0.3\linewidth]{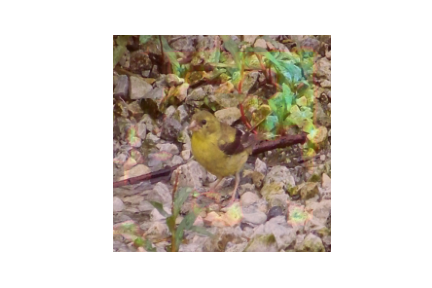}
    \includegraphics[trim={3.6cm 1cm 3.6cm 1cm},clip,width=0.3\linewidth]{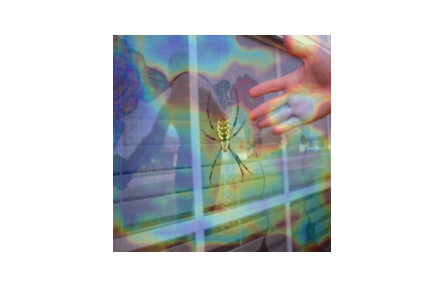}
    \\
     \includegraphics[trim={3.6cm 1cm 3.6cm 1cm},clip,width=0.3\linewidth]{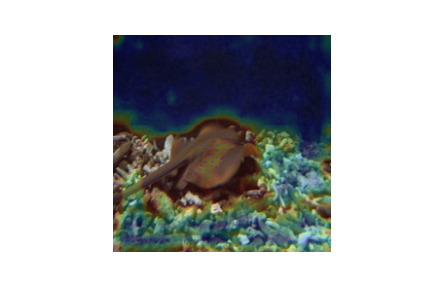}
    \includegraphics[trim={3.6cm 1cm 3.6cm 1cm},clip,width=0.3\linewidth]{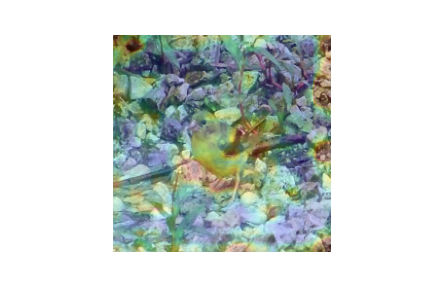}
    \includegraphics[trim={3.6cm 1cm 3.6cm 1cm},clip,width=0.3\linewidth]{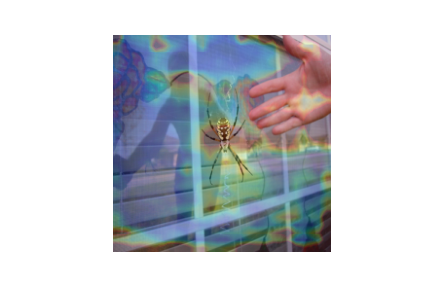}
    \\
    \includegraphics[trim={3.6cm 1cm 3.6cm 1cm},clip,width=0.3\linewidth]{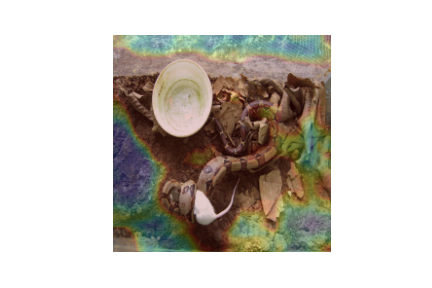}
    \includegraphics[trim={3.6cm 1cm 3.6cm 1cm},clip,width=0.3\linewidth]{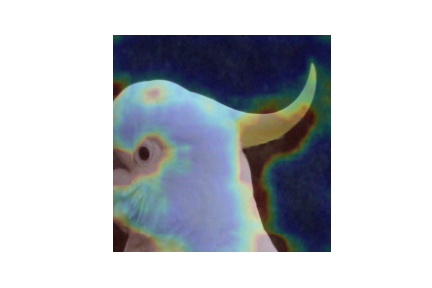}
    \includegraphics[trim={3.6cm 1cm 3.6cm 1cm},clip,width=0.3\linewidth]{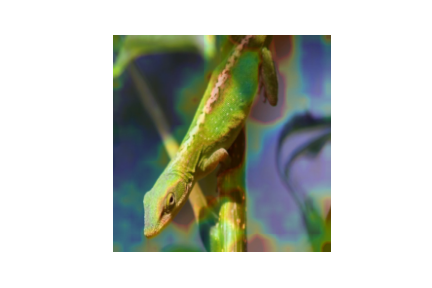}
    \\
    \includegraphics[trim={3.6cm 1cm 3.6cm 1cm},clip,width=0.3\linewidth]{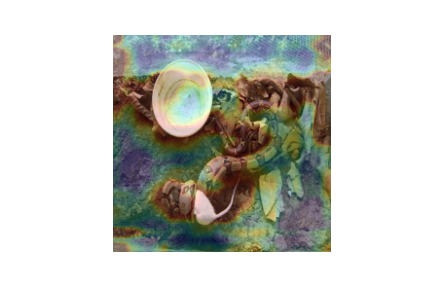}
    \includegraphics[trim={3.6cm 1cm 3.6cm 1cm},clip,width=0.3\linewidth]{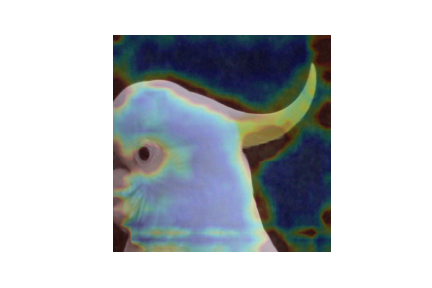}
    \includegraphics[trim={3.6cm 1cm 3.6cm 1cm},clip,width=0.3\linewidth]{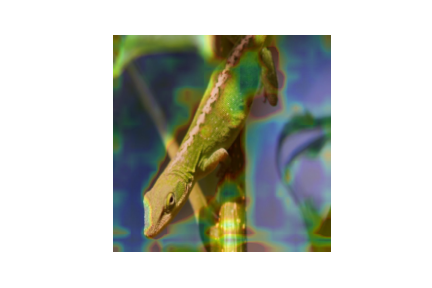}
    \textbf{}
    
    \caption{Attribution maps from ImageNet100 computed using LayerCAM: \textbf{Row 1,3:} model trained without 
    GradMix; \textbf{Row 2,4:} model trained with GradMix.
    Models with GradMix demonstrate better object focus in data.}
    \label{app-fig-imagenet100}
\end{figure}

We present in \cref{app-fig-imagenet100} the attribution maps of self-supervised models trained on the ImageNet100 dataset, as discussed in \cref{subsec-exe-generalization}, to demonstrate the effectiveness of \emph{GradMix}. By comparing the attribution maps from models trained with and without \emph{GradMix}, we observe that \emph{GradMix} enhances the models' ability to capture objects more effectively. Specifically, models trained with \emph{GradMix} exhibit a reduced focus on background regions and concentrate more precisely on foreground objects, even in a self-supervised learning setting. For example, in the second pair in \cref{app-fig-imagenet100} (second images of row 1 and 2), the goldfish are more precisely captured when with \emph{GradMix}. And in the fifth pair (second images of row 3 and 4), the model learns almost just the background when without \emph{GradMix}.

\section{More Visualizations and Analysis}  \label{app-more-visualizations}
\begin{figure}[ht]
\centering
     \includegraphics[trim={3.6cm 1cm 3.6cm 1cm},clip,width=0.2\linewidth]{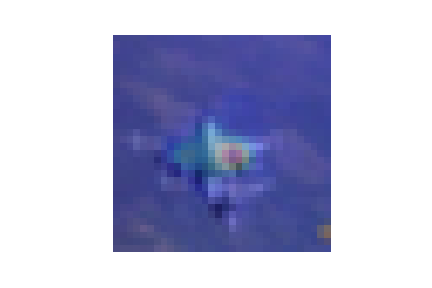}
     \includegraphics[trim={3.6cm 1cm 3.6cm 1cm},clip,width=0.2\linewidth]{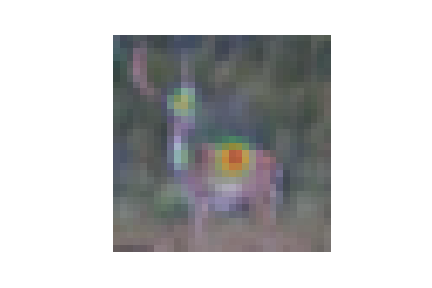}
     \includegraphics[trim={3.6cm 1cm 3.6cm 1cm},clip,width=0.2\linewidth]{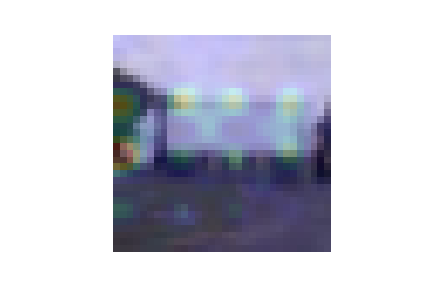}
     \includegraphics[trim={3.6cm 1cm 3.6cm 1cm},clip,width=0.2\linewidth]{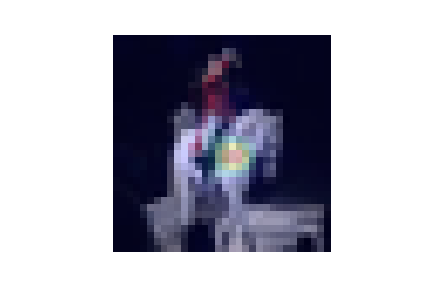}
    \\
    \includegraphics[trim={3.6cm 1cm 3.6cm 1cm},clip,width=0.2\linewidth]{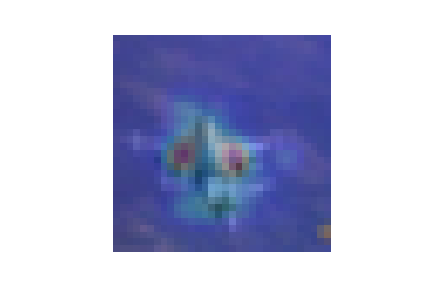}
    \includegraphics[trim={3.6cm 1cm 3.6cm 1cm},clip,width=0.2\linewidth]{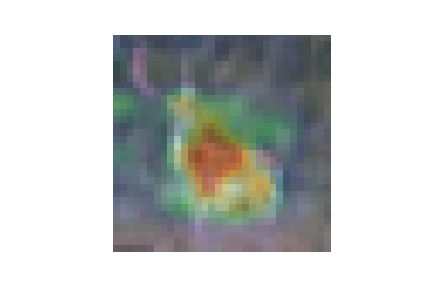}
    \includegraphics[trim={3.6cm 1cm 3.6cm 1cm},clip,width=0.2\linewidth]{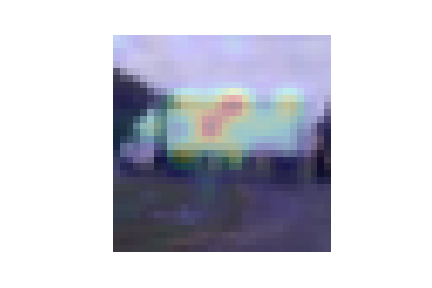}
    \includegraphics[trim={3.6cm 1cm 3.6cm 1cm},clip,width=0.2\linewidth]{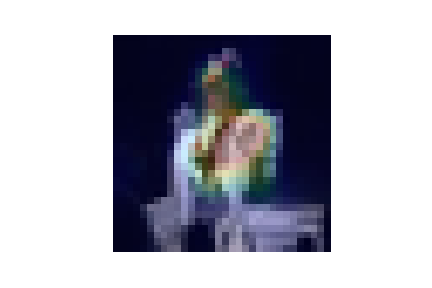}
    \\
    \includegraphics[trim={3.6cm 1cm 3.6cm 1cm},clip,width=0.2\linewidth]{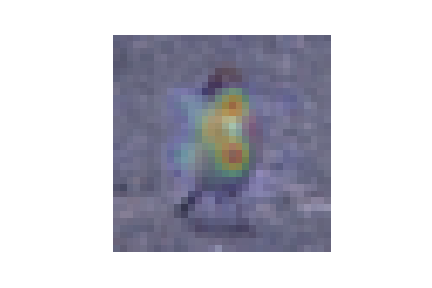}
    \includegraphics[trim={3.6cm 1cm 3.6cm 1cm},clip,width=0.2\linewidth]{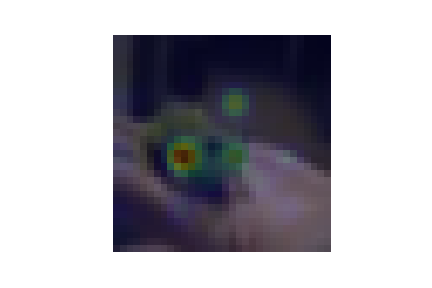}
    \includegraphics[trim={3.6cm 1cm 3.6cm 1cm},clip,width=0.2\linewidth]{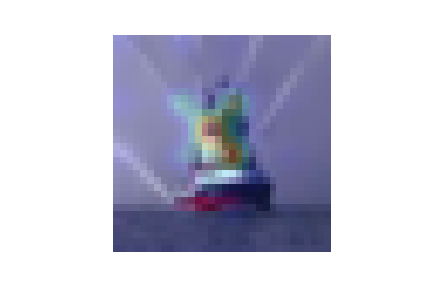}
    \includegraphics[trim={3.6cm 1cm 3.6cm 1cm},clip,width=0.2\linewidth]{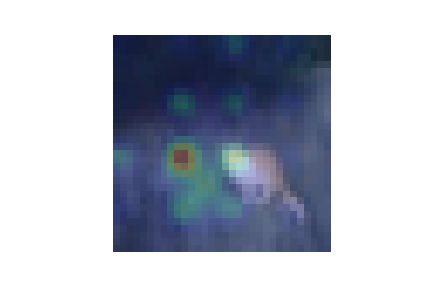}
    \\
    \includegraphics[trim={3.6cm 1cm 3.6cm 1cm},clip,width=0.2\linewidth]{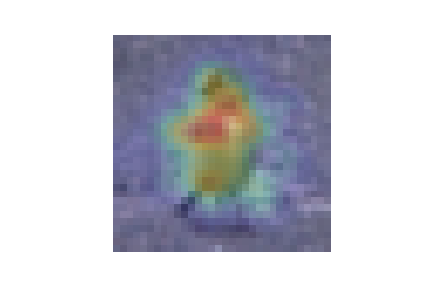}
    \includegraphics[trim={3.6cm 1cm 3.6cm 1cm},clip,width=0.2\linewidth]{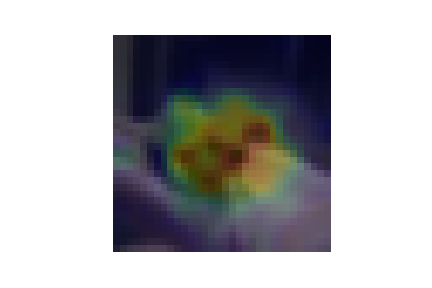}
    \includegraphics[trim={3.6cm 1cm 3.6cm 1cm},clip,width=0.2\linewidth]{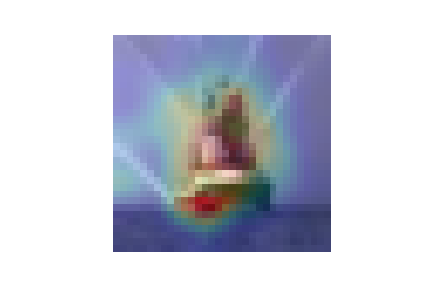}
    \includegraphics[trim={3.6cm 1cm 3.6cm 1cm},clip,width=0.2\linewidth]{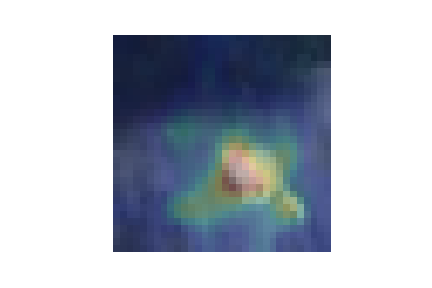}
    \\
    \includegraphics[trim={3.6cm 1cm 3.6cm 1cm},clip,width=0.2\linewidth]{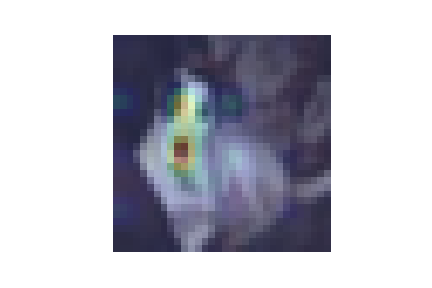}
    \includegraphics[trim={3.6cm 1cm 3.6cm 1cm},clip,width=0.2\linewidth]{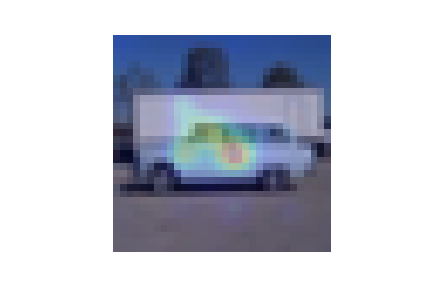}
    \includegraphics[trim={3.6cm 1cm 3.6cm 1cm},clip,width=0.2\linewidth]{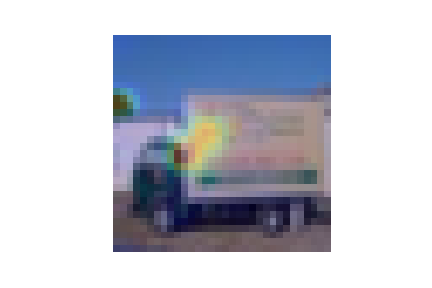}
    \includegraphics[trim={3.6cm 1cm 3.6cm 1cm},clip,width=0.2\linewidth]{images/deer_plain/179.png}
    \\
    \includegraphics[trim={3.6cm 1cm 3.6cm 1cm},clip,width=0.2\linewidth]{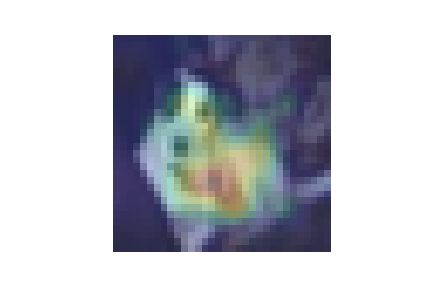}
    \includegraphics[trim={3.6cm 1cm 3.6cm 1cm},clip,width=0.2\linewidth]{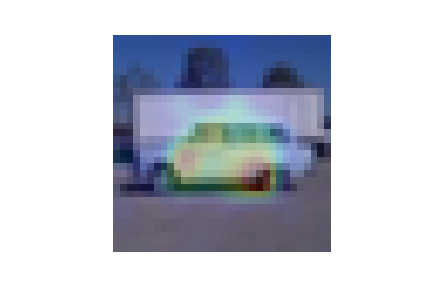}
    \includegraphics[trim={3.6cm 1cm 3.6cm 1cm},clip,width=0.2\linewidth]{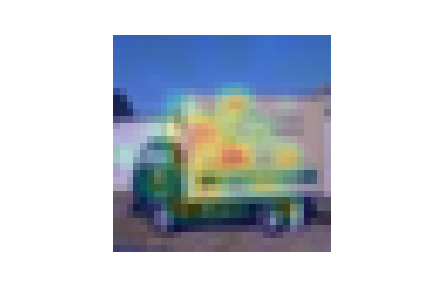}
    \includegraphics[trim={3.6cm 1cm 3.6cm 1cm},clip,width=0.2\linewidth]{images/deer_grad/179.png}
    \\
    \includegraphics[trim={3.6cm 1cm 3.6cm 1cm},clip,width=0.2\linewidth]{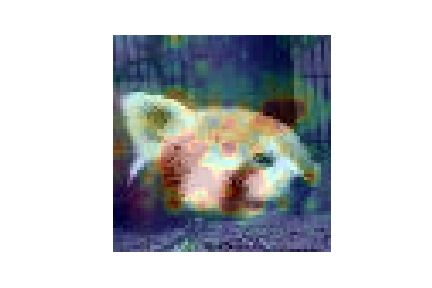}
    \includegraphics[trim={3.6cm 1cm 3.6cm 1cm},clip,width=0.2\linewidth]{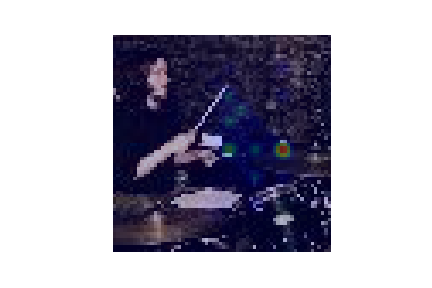}
    \includegraphics[trim={3.6cm 1cm 3.6cm 1cm},clip,width=0.2\linewidth]{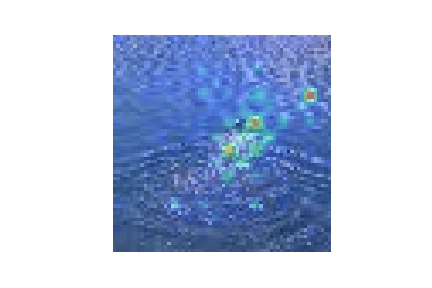}
    \includegraphics[trim={3.6cm 1cm 3.6cm 1cm},clip,width=0.2\linewidth]{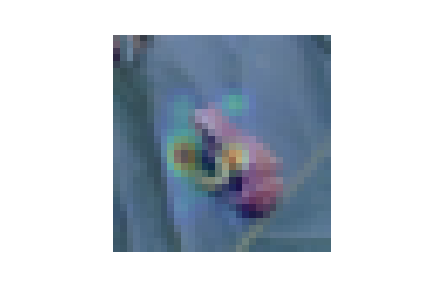}
    \\
    \includegraphics[trim={3.6cm 1cm 3.6cm 1cm},clip,width=0.2\linewidth]{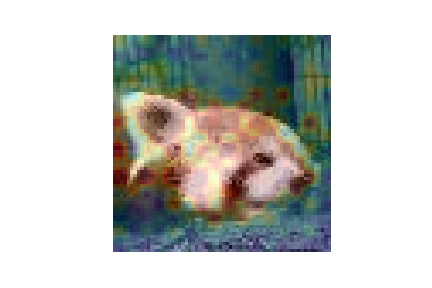}
    \includegraphics[trim={3.6cm 1cm 3.6cm 1cm},clip,width=0.2\linewidth]{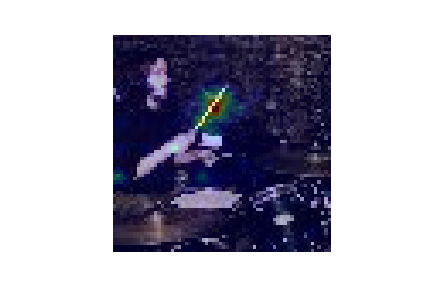}
    \includegraphics[trim={3.6cm 1cm 3.6cm 1cm},clip,width=0.2\linewidth]{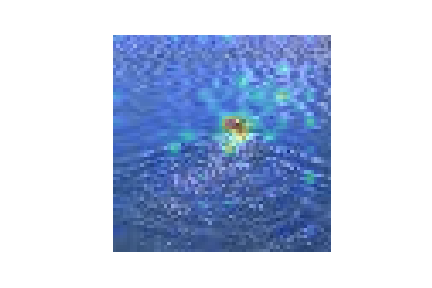}
    \includegraphics[trim={3.6cm 1cm 3.6cm 1cm},clip,width=0.2\linewidth]{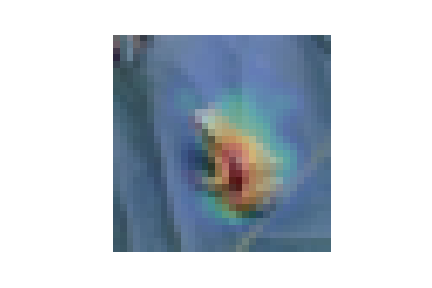}
    
    \caption{Attribution maps from CIFAR10 and TinyImageNet datasets computed using LayerCAM: \textbf{Row 1,3,5,7:} model trained without 
    GradMix; \textbf{Row 2,4,6,8:} model trained with GradMix.
    Models with GradMix demonstrate larger activated areas and better object focus in data.}
    \label{app-fig-visualization}
\end{figure}

Following section \ref{sec-analysis-discussion}, the attribution maps of more samples are demonstrated in Figure \ref{app-fig-visualization}. Aligned with \ref{sec-analysis-discussion}, models with GradMix focus on a broader range of regions in the data. 

\section{Introduction on FMs for OOD Detection and Discussion}   \label{app-fm-ood}

Foundation models have increasingly been applied to out of distribution (OOD) detection problems in recent years \cite{ming2022delving, li2024learning, miyai2024locoop, wang2023clipn}, with particular attention to vision-language models like CLIP \cite{radford2021learning}. We introduce the representative works in the following text:

\noindent\textbf{Maximum Concept Matching (MCM)} \cite{ming2022delving}
proposes constructing prototype representations for in distribution (ID) data using CLIP's text encoder outputs. These prototypes are generated by prompting the encoder with common textual descriptions of the class names, such as "This is a photo of [class name]." The visual representations of test samples are then compared to the prototypes using the proposed Maximum Concept Matching score, which is defined as the softmax of the cosine similarity between the visual and prototype representations.

\noindent\textbf{NegPrompt} \cite{li2024learning} takes a different approach from MCM, which relies on common prompts to obtain text representations of ID classes. Instead, NegPrompt learns negative prompts for OOD data, where each negative prompt represents a negative connotation of an ID class. The method assumes that the representations of negative prompts are distinct from those of positive prompts. Prediction scores are computed based on the similarities between the test image representations and both the positive and negative prompts.

\noindent\textbf{CLIPN} \cite{wang2023clipn} trains as well negative prompts that integrate negation semantics of each prompt for ID data and a negative text encoder in CLIP. Similar to NegPrompt, the learning objective ensures that the image representations of ID data are close to their corresponding positive text representations while being far from their associated negative prompts. 
The likelihood of a test sample being out of distribution is then computed based on the similarities between the image representations and both the positive text representations and the text representations of the negative prompts generated by the negative text encoder.

\noindent\textbf{LoCoOp} \cite{miyai2024locoop} enhances text prompt learning by excluding class-irrelevant local regions in ID data. During training, regions within the original ID samples with low similarity between their visual representations and the text representations of the corresponding ID classes are removed. This process ensures that the learned text prompts are more focused on class-relevant features. During inference, the MCM score that is based on the similarities between visual and text representations is utilized to identify OOD samples.

\vspace{5 mm}

As discussed in \ref{sec-conclusion}, all the methods introduced above operate on the fundamental premise that image and text representations are well-aligned in the latent space. These approaches focus on learning precise prompts for in distribution data, and in some cases for out of distribution data as well, to enhance the discrimination between ID and OOD test samples. This discrimination is achieved by evaluating the similarities between the visual representations of the test samples and the text representations of the prompts. In other words, the text representations of the prompts are utilized to model the dataset effectively.

\end{document}